\documentclass[lettersize,journal]{IEEEtran}
\usepackage{amsmath,amsfonts}
\usepackage{algorithmic}
\usepackage{algorithm}
\usepackage{array}
\usepackage[caption=false,font=normalsize,labelfont=sf,textfont=sf]{subfig}
\usepackage{textcomp}
\usepackage{stfloats}
\usepackage{url}
\usepackage{verbatim}
\usepackage{graphicx}
\usepackage{cite}
\usepackage{booktabs}       
\usepackage{multirow}
\usepackage{diagbox}
\usepackage{wrapfig}
\usepackage{colortbl}
\usepackage{color}
\usepackage{soul}
\definecolor{orange}{rgb}{0.910,0.631,0.580}
\definecolor{blue}{rgb}{0.580,0.733,0.910}

\newcommand{\etal}{\textit{et al.}}

\begin{document}

\title{Monocular Road Planar Parallax Estimation}

\author{Haobo~Yuan,
        Teng~Chen,
        Wei~Sui,
        Jiafeng~Xie,
        Lefei~Zhang,~\IEEEmembership{Senior Member, IEEE}\\
        Yuan~Li,
        and~Qian~Zhang
\thanks{This work was supported by the National Natural Science Foundation of China under Grant 62122060 and the Special Fund of Hubei Luojia Laboratory under Grant 220100014. (Corresponding Author: Lefei Zhang; Equal Contribution: Haobo Yuan and Teng Chen)}
\thanks{Haobo Yuan and Lefei Zhang are with National Engineering Research Center for Multimedia Software, School of Computer Science, Wuhan University, Wuhan, 430072, China. Lefei Zhang is also with Hubei Luojia Laboratory, Wuhan, China. Email: \{yuanhaobo, zhanglefei\}@whu.edu.cn}
\thanks{Teng Chen, Wei Sui, Jiafeng Xie, and Qian Zhang are with Horizon Robotics, Haidian, Beijing, 100083, China. Email: \{teng.chen, wei.sui, jiafeng.xie, qian01.zhang\}@horizon.ai}
\thanks{Yuan Li is with Google Research, Mountain View, CA 94043, USA. Email: liyu@google.com}
\thanks{Digital Object Identifier: xx.xxxx/xxxx}}

\markboth{Journal of \LaTeX\ Class Files,~Vol.~14, No.~8, August~2021}%
{Shell \MakeLowercase{\textit{et al.}}: A Sample Article Using IEEEtran.cls for IEEE Journals}

\IEEEpubid{0000--0000/00\$00.00~\copyright~2021}

\maketitle

\begin{abstract}
Estimating the 3D structure of the drivable surface and surrounding environment is a crucial task for assisted and autonomous driving. It is commonly solved either by using 3D sensors such as LiDAR or directly predicting the depth of points via deep learning. However, the former is expensive, and the latter lacks the use of geometry information for the scene. In this paper, instead of following existing methodologies, we propose Road Planar Parallax Attention Network (RPANet), a new deep neural network for 3D sensing from monocular image sequences based on planar parallax, which takes full advantage of the omnipresent road plane geometry in driving scenes. RPANet takes a pair of images aligned by the homography of the road plane as input and outputs a $\gamma$ map (the ratio of height to depth) for 3D reconstruction. The $\gamma$ map has the potential to construct a two-dimensional transformation between two consecutive frames. It implies planar parallax and can be combined with the road plane serving as a reference to estimate the 3D structure by warping the consecutive frames. Furthermore, we introduce a novel cross-attention module to make the network better perceive the displacements caused by planar parallax. To verify the effectiveness of our method, we sample data from the Waymo Open Dataset and construct annotations related to planar parallax. Comprehensive experiments are conducted on the sampled dataset to demonstrate the 3D reconstruction accuracy of our approach in challenging scenarios.
\end{abstract}


\section{Introduction}
\IEEEPARstart{3}{D} reconstruction of road environment~\cite{asai20083d,chen2015fast,gao20183d} has received increasing attention in the assisted and autonomous driving field~\cite{badue2020self,schoettle2014survey} as it is crucial for obstacle detection~\cite{lu2020autonomous}, distance measurement~\cite{xu2020aanet,zhang2019ga}, and road condition recognition~\cite{fujita2020fine}, etc. 
Existing methods commonly exploit 3D sensors such as LiDAR or adopt vision-based 3D reconstruction algorithms. 3D sensors \cite{zhang2014loam} can provide reasonably accurate 3D information, but the high price, the sparse nature, along with the concerns about reliability limit their potential in mass production. In contrast, vision-based methods such as Structure from Motion (SfM)~\cite{schonberger2016structure} are low-cost yet suffer from various conditions such as weakly-textured regions, lighting variation, and rapid movement. Besides, these methods usually require laborious manual parameter tuning to guarantee good performance. Recently, deep learning-based methods have been applied to 3D road reconstruction \cite{godard2019digging,yin2018geonet,tishchenko2020self} and have shown promising performance. In these methods, the learning-based depth estimation tasks can be summarized as directly regressing depth numerical values from image pixels using convolutional neural networks (CNNs) trained in a supervised~\cite{eigen2014depth} or unsupervised manner~\cite{zhou2017unsupervised}. 
However, simply applying deep neural networks to depth estimation and considering it as a per-pixel regression task means a lack of use of scene geometry information.

\IEEEpubidadjcol

In this paper, we would like to bring attention to a family of methods utilizing planar parallax geometry for 3D reconstruction~ \cite{irani1996parallax,irani1997recovery,kumar1994direct,shashua1994relative}. Planar parallax was first proposed in the 1990s used for planar motion modeling. The core idea behind it is that the 3D structure is strongly related to the residual image displacements caused by homography warping between two views. Although planar parallax-based methods require a "plane" as a reference, which may be hard to find in some scenes, 3D road reconstruction is naturally a good application of planar parallax as the ground plane serves perfectly as the reference plane. However, they are susceptible to image noise and only suitable for rigid scenes, which prevents them from being widely adopted~\cite{irani2002direct}. In the real-world datasets on the way, such as Waymo Open Dataset~\cite{sun2020scalability}, the traditional planar parallax method may fail according to our experiments.
 
To overcome the drawbacks of traditional methods, we design a deep neural network named RPANet to estimate dense planar parallax. RPANet utilizes a novel cross-attention mechanism, takes a pair of images aligned by road plane homography as input, and outputs a pixel-wise $\gamma$ map that represents the pixel-wise ratio of height to depth. A photometric loss could then be applied to train the network since the planar parallax is a \textit{de facto} bridge between the homography-aligned images. Note that the photometric loss relies on the residual flow, which is in the realm of traditional planar parallax geometry. Leveraging both traditional geometry and deep learning with the derived geometry formulas, our method benefits from the robustness of deep learning and the interpretability of geometry-based algorithms.

As there is no publicly available dataset for planar parallax estimation on the road, we build a Road Planar Parallax Dataset (RP2-Waymo) based on the Waymo Open Dataset~\cite{sun2020scalability} to train and evaluate our method. We sample data from the Waymo Open Dataset for its diversity in scenes, seasons, and time of day as well as the excellent synchronization between LiDAR and cameras. Beyond the original samples, we estimate the road plane from the LiDAR points, with which both the homography matrix and the ground truth $\gamma$ can be computed. Sparse ground truth of depth and height is generated by projecting LiDAR points to images. Except for the depth and height, we construct the high-precision road homography, which warps the input image pairs. Thanks to the homography matrix, we can get the image pairs with the aligned road, which is the input of our proposed RPANet. The main contributions of our work are summarized as follows:
\begin{itemize}
    \item Inspired by the traditional planar parallax geometry, we propose to take the omnipresent road as the reference for 3D structure estimation with deep learning. The predicted planar parallax can be used to reconstruct depth and height of each pixel as well as boost the learning.
    \item Motivated by the attention mechanism in the stereo-matching, we propose a novel deep neural network called RPANet, which leverages a novel cross-attention module. The cross-attention module can be utilized to find the matching relationship between two images easily and is conducive to predicting the planar parallax.
    \item To validate our proposed method, we build the RP2-Waymo dataset based on the Waymo Open Dataset~\cite{sun2020scalability} for planar parallax estimation. Extensive experiments are conducted on this dataset, and the results demonstrate that our method can recover accurate 3D road surface structures.
\end{itemize}

\section{Related Work}
\paragraph{Planar Parallax}
The planar parallax model is first proposed in \cite{sawhney19943d, shashua1994relative} to derive a 3D structure relative to a planar surface. They demonstrate that by leveraging the planar parallax model to remove the camera rotation, the reconstruction becomes more accurate and stable. Inspired by \cite{sawhney19943d, shashua1994relative}. Kumar \etal~\cite{kumar1994direct} propose a method applying the planar parallax model to estimate the height in aerial images. Observing that the depth of all points in aerial images is nearly the same, they eliminate the depth factor in the parallax equation by calculating the normal of the ground plane. Irani \etal~\cite{irani2002direct} further extend the two frames algorithms of \cite{sawhney19943d, shashua1994relative} to multiple frames and improve the robustness of 3D structure reconstruction. However, most traditional planar parallax algorithms need to obtain accurate correspondence in advance, therefore susceptible to noise. Furthermore, traditional (non-learning-based) planar parallax algorithms also suffer from low speed. haney \etal~\cite{chaney2019learning} train a deep neural network constrained by planar parallax for the the event-based camera. A concurrent work~\cite{xing2022joint} also uses planar parallax geometry but focusing more on depth estimation. Our aim in this paper is to utilize deep learning to estimate planar parallax, which can effectively determine the depth and height of each pixel, to estimate the structure of a driving scene.

\paragraph{Learning-based 3D Structure Estimation}
One of the earliest works trying to estimate 3D structure from a 2D image by the neural network is proposed by Eigen \etal~\cite{eigen2014depth}. They directly regress the depth information from a single image. 
Except for their attempt to predict the depth map to estimate 3D structure, other works try to estimate the 3D structure by predicting the point cloud~\cite{fan2017point}, voxel~\cite{choy20163d}, mesh~\cite{wang2018pixel2mesh}, or implicit function~\cite{saito2019pifu}. However, although these works~\cite{fan2017point,choy20163d,wang2018pixel2mesh,saito2019pifu} have the potential to predict the whole 3D models, even including the backside, they mainly focus on the single object 3D reconstruction rather than scene 3D reconstruction.
At the same time, especially for the autonomous driving use case, geometric constraints are introduced to boost the prediction of dense depth. E.g., \cite{garg2016unsupervised, godard2017unsupervised, zhou2017unsupervised, godard2019digging} use 3D geometry constraints to train the depth estimation networks by re-projecting the depth map, while \cite{zbontar2015computing, kendall2017end, chang2018pyramid, guo2019group} construct constraints through stereo geometry.
The application of different 3D geometric constraints above helps reduce the difficulty of training and further improves the accuracy of 3D structure estimation. Distinct from the previous works, we apply a novel geometric constraint based on planar parallax.

\begin{figure*}
  \subfloat[]{\includegraphics[width=.49\textwidth]{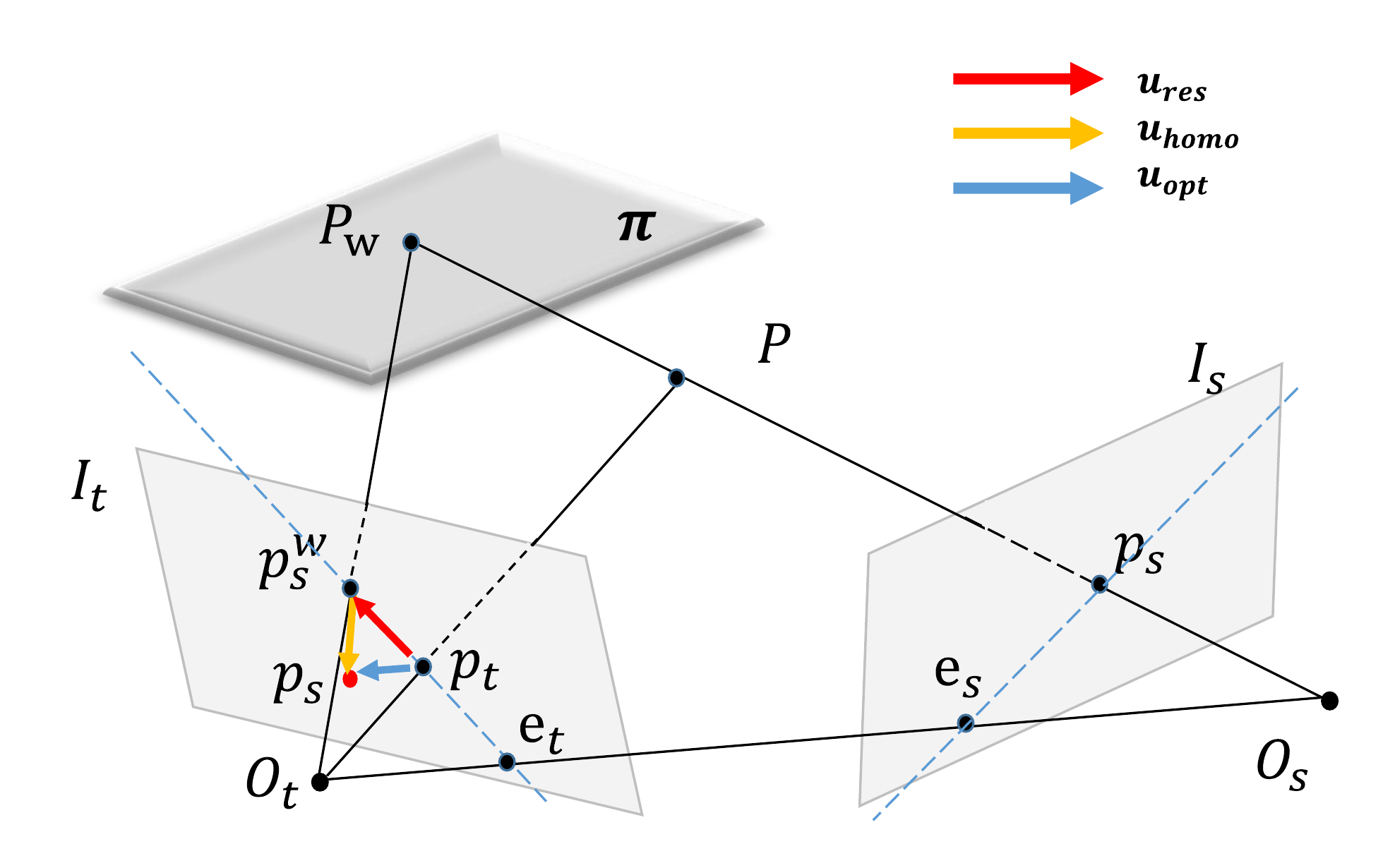}%
    \label{fig:geoa}}
    \hfil
    \subfloat[]{\includegraphics[width=.49\textwidth]{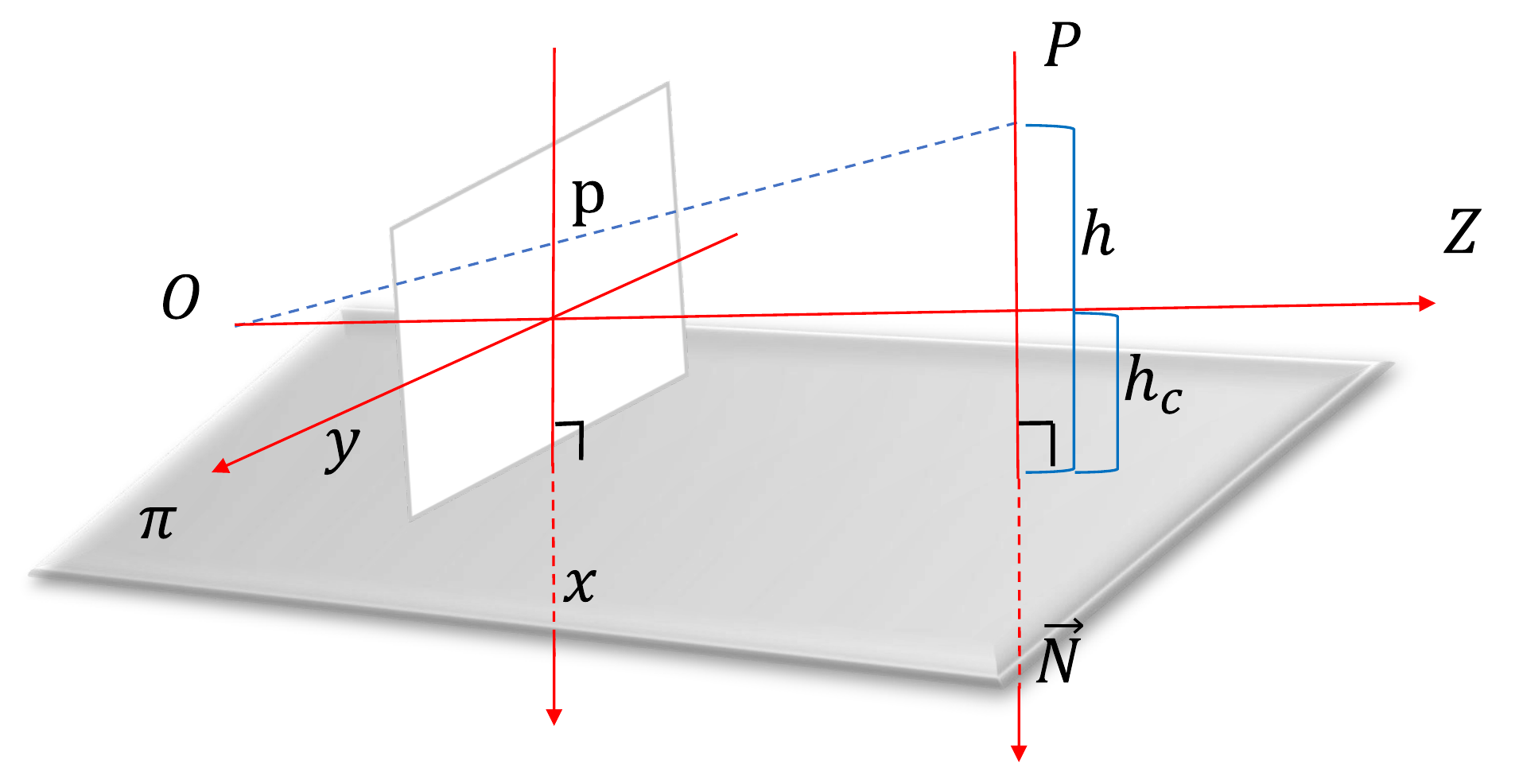}%
    \label{fig:geob}}
\caption{(a) The planar parallax geometry. (b) The illustration of 3D reconstruction from $\gamma$.} \label{fig:geo}
\end{figure*}

\paragraph{Visual Transformer}
Transformer~\cite{vaswani2017attention} is first used in the Natural Language Processing (NLP) tasks and origins from the self-attention mechanism. It shows extraordinary performance on many NLP tasks~\cite{han2020survey}. Inspired by the self-attention mechanism, many researchers explore a similar mechanism to solve computer vision tasks, such as classification~\cite{dosovitskiy2021image}, object detection~\cite{carion2020end, zhu2021deformable} and image generation~\cite{parmar2018image}. In general, the attention mechanism can enhance the global perception of neural networks beyond CNNs~\cite{dosovitskiy2021image, liu2021Swin}. However, the self-attention mechanism needs lots of computational resources. They often use patch-based~\cite{dosovitskiy2021image} or axial-based~\cite{ho2019axial} strategies to reduce the computational overhead. In the field of 3D structure estimation, thanks to the epipolar constraint, some stereo match schemes~\cite{wang2020parallax, li2020revisiting} also apply attention-based mechanisms to build their neural networks without extremely high computational costs. In our proposed method, we also apply a proposed cross-attention mechanism to model the displacement caused by the homography transformation.

\section{Our Approach}
Instead of predicting depth directly through deep neural network such as in \cite{godard2017unsupervised, zhou2017unsupervised, godard2019digging, guizilini20203d}, our proposed method estimates a $\gamma$ map, which is denoted as
\begin{equation} \label{eq:depth}
    \gamma = \frac{h}{d}.
\end{equation}
The $\gamma$ map represents the ratio of height $h$ to depth $d$ for each pixel from two consecutive images $I_s$ and $I_t$. 
The pair of input images would be aligned twice, first by the road plane homography and then by the residual flow generated from the $\gamma$ map.
After the final alignment, the static region between two images should be well aligned, hence photometric error can be calculated.

\subsection{Planar Parallax Geometry} \label{sec:planar_parallax}
\paragraph{Planar Parallax.}
 A reference plane is denoted as $\mathbf{\pi}$ in 3D space and $\mathbf{P}$ is a point off $\mathbf{\pi}$.  The point $\mathbf{P}$ is observed by two camera views whose optical centers are represented as $\mathbf{O}_t$ and $\mathbf{O}_s$. $\mathbf{p_t}$ and $\mathbf{p}_{s}$ are the re-projections on image $I_t$ and $I_s$ respectively. Supposing $\mathbf{P}^{w}$ is the intersection of the ray $\overrightarrow{\mathbf{O}_{s}\mathbf{P}}$ and $\mathbf{\pi}$, we can obtain its re-projection in camera ${\mathbf{O}_{t}}$ denoted as $\mathbf{p}_{s}^{w}$ by the homography $\mathbf{H}_{s\to t}$ (See Fig.~\ref{fig:geoa} for details).
Based on this, the relationship of $\mathbf{p}_{t}$, $\mathbf{p}_{s}$, and $\mathbf{p}_{t}^{w}$
can be represented as follows:
\begin{equation}\label{eq:def}
\begin{aligned}
  \mathbf{u}_{homo} &= \mathbf{p}_{s} - \mathbf{p}_{s}^{w} ,\\
  \mathbf{u}_{res} &= \mathbf{p}_{s}^{w} - \mathbf{p}_{t} ,\\
  \mathbf{u}_{opt} &= \mathbf{p}_{s} - \mathbf{p}_{t} = \mathbf{u}_{homo} + \mathbf{u}_{res},
\end{aligned}
\end{equation}
where $\mathbf{u}_{homo}$ is the displacement caused by homography while $\mathbf{u}_{res}$ is the residual flow. The residual flow represents the displacement of corresponding pixels between a pair of images already aligned by the road homography. We hold that $\mathbf{u}_{opt}$ is a de facto bridge between traditional 3D geometry ($\mathbf{u}_{homo}$) and deep neural network ($\mathbf{u}_{res}$ derived from $\gamma$).

In Fig.~\ref{fig:geoa}, following~\cite{zhou2017unsupervised} each 3D point $\mathbf{P'}$ in the source camera coordinate system denoted by $\mathbf{O}_s$ could be transformed to the target camera coordinate system denoted by $\mathbf{O}_t$ by a rigid transformation 
\begin{equation}\label{eq:d1}
{\mathbf{P}} = \mathbf{R} {\mathbf{P'}} + {\mathbf{T}},
\end{equation}
where ${\mathbf{T}} = (\mathbf{T}_x, \mathbf{T}_y, \mathbf{T}_z)$ is the translation vector in $\mathbf{O}_t$, and $\mathbf{R}$ is the rotation matrix from $\mathbf{O}_s$ to $\mathbf{O}_t$.

Except for the depth, we need to introduce the height of each 3D point in our proposed method. Given an arbitrary $\mathbf{P}=(X, Y, Z)$ in camera coordinate system, where $Z$ is the depth, its height could be expressed as 
\begin{equation}\label{eq:1}
   h = h_c - \vec{\mathbf{N}}^T\mathbf{P}.
\end{equation}
where $\vec{\mathbf{N}}$ is the normal of $\mathbf{\pi}$, and $h_c$ is the height of camera to plane $\mathbf{\pi}$. Then, we have
\begin{equation}\label{eq:d4}
   \frac{h + \vec{\mathbf{N}}^T {\mathbf{P}} }{h_c} = 1 .
\end{equation}
Multiplying ${\mathbf{T}}$ by Eqn.~\ref{eq:d4} we could obtain 
\begin{equation}\label{eq:d5}
\begin{aligned}
   {\mathbf{P}} &= \mathbf{R} {\mathbf{P}}' + {\mathbf{T}} \frac{h-\vec{\mathbf{N}}^T {\mathbf{P}}'}{h_c} \\ 
   &= (\mathbf{R} + \frac{{\mathbf{T}} \vec{\mathbf{N}}^T}{h_c}){\mathbf{P}}' + \frac{h}{h_c} {\mathbf{T}}.
\end{aligned}
\end{equation}

With $\mathbf{t} =\mathbf{K} \mathbf{T} = (t_x, t_y, t_z)^T$, $\mathbf{p}= \frac{\mathbf{K}}{Z} \mathbf{P}= (x, y, 1)^T$,
$\mathbf{p}' = \frac{\mathbf{K}'}{Z'} \mathbf{P}' = (x', y', 1)^T$, we obtain 
\begin{equation}\label{eq:d6}
\begin{aligned}
   Z\mathbf{K}^{-1}{\mathbf{p}} = (\mathbf{R} + \frac{{\mathbf{T}} \vec{\mathbf{N}}^T}{h_c})Z'\mathbf{K}^{-1} {\mathbf{p}}' + \frac{h}{h_c }{\mathbf{T}}.
\end{aligned}
\end{equation}
We multiply Eqn.~\ref{eq:d6} by $\frac{\mathbf{K}}{Z'}$ on both sides and obtain
\begin{equation}\label{eq:d7}
\begin{aligned}
  \frac{Z}{Z'}{\mathbf{p}} &= \mathbf{K}(\mathbf{R} + \frac{{\mathbf{T}} \vec{\mathbf{N}}^T}{h_c})\mathbf{K}^{-1} {\mathbf{p}}' + \frac{h}{h_c Z'}{\mathbf{t}} \\
  &= \mathbf{H} {\mathbf{p}}' + \frac{h}{h_c Z'}{\mathbf{t}},
\end{aligned}
\end{equation}
where $\mathbf{H}=\mathbf{K}(\mathbf{R} + \frac{{\mathbf{T}} \vec{\mathbf{N}}^T}{h_c})\mathbf{K}^{-1}$ represents the homography matrix between the two images. 
 Eqn.~\ref{eq:d7} could be reformulated as
\begin{equation}\label{eq:d8}
\begin{aligned}
{\mathbf{p}}  = \frac{\mathbf{H} {\mathbf{p}}' + \frac{h}{h_c Z'}{\mathbf{t}}}{\frac{Z}{Z'}}.
\end{aligned}
\end{equation}
The z-axis of ${\mathbf{p}}$ and ${\mathbf{p}}'$ is 1. 
Only the third row of $\mathbf{H}$ and ${\mathbf{t}}$ contains the information of z-axis. 
To obtain ${\mathbf{p}} =(x, y, 1)$, we apply $\frac{Z}{Z'}$ to normalize the z-axis of $\mathbf{H} {\mathbf{p}'} + \frac{h}{h_c Z'}{\mathbf{t}}$.  
Notice that we could get the derivation
\begin{equation}\label{eq:d9}
\frac{Z}{Z'} =\mathbf{H}_3 {\mathbf{p}}' +\frac{h \mathbf{T}_z}{h_c Z'},
\end{equation}
where $\mathbf{H}_3$ and $\mathbf{T}_z$ are third component of $\mathbf{H}$ and $\mathbf{T}$ respectively.
By substituting $\frac{Z}{Z'}$, we obtain 
\begin{equation}\label{eq:d10}
\begin{aligned}
{\mathbf{p}}  &= \frac{\mathbf{H} {\mathbf{p}}' + \frac{h}{h_c Z'}{\mathbf{t}}}{\mathbf{H}_3 {\mathbf{p}}' +\frac{h \mathbf{T}_z}{h_c Z'}} 
\\
&= \frac{\mathbf{H} {\mathbf{p}}'}{\mathbf{H}_3 {\mathbf{p}}'}-\frac{\mathbf{H} {\mathbf{p}}'}{\mathbf{H}_3 {\mathbf{p}'}} +\frac{\mathbf{H} {\mathbf{p}}' + \frac{h}{h_c Z'}{\mathbf{t}}}{\mathbf{H}_3 {\mathbf{p}}' +\frac{h \mathbf{T}_z}{h_c Z'}} 
\\
&=\frac{\mathbf{H} {\mathbf{p}}'}{\mathbf{H}_3 {\mathbf{p}}'} - \frac{\frac{h \mathbf{T}_z}{h_c Z'}}{\mathbf{H}_3 {\mathbf{p}}' + \frac{h \mathbf{T}_z}{h_c Z'}}\frac{\mathbf{H} {\mathbf{p}}'}{\mathbf{H}_3 {\mathbf{p}}'} +\frac{\frac{h}{h_c Z'}}{\mathbf{H}_3 {\mathbf{p}}'+\frac{h \mathbf{T}_z}{h_c Z'}} {\mathbf{t}} 
\\
&=\frac{\mathbf{H} {\mathbf{p}}'}{\mathbf{H}_3 {\mathbf{p}}'} - \frac{h \mathbf{T}_z}{Z h_c}\frac{\mathbf{H} {\mathbf{p}}'}{\mathbf{H}_3 {\mathbf{p}}'} + \frac{h}{Z h_c} {\mathbf{t}}.
\end{aligned}
\end{equation}
When $\mathbf{T}_z = 0$, Eqn.~\ref{eq:d10} becomes
\begin{equation}\label{eq:d11}
\begin{aligned}
 \mathbf{p} = \frac{\mathbf{H} {\mathbf{p}}'}{\mathbf{H}_3 {\mathbf{p}}'} + \frac{h}{Z h_c} {\mathbf{t}}.
\end{aligned}
\end{equation}
When $\mathbf{T}_z \neq 0$, we obtain 
\begin{equation}\label{eq:d12}
\begin{aligned}
{\mathbf{p}} = \frac{\mathbf{H} {\mathbf{p}}'}{\mathbf{H}_3 {\mathbf{p}}'} - \frac{h \mathbf{T}_z}{Z h_c} (\frac{\mathbf{H} {\mathbf{p}}'}{\mathbf{H}_3 {\mathbf{p}}'}-{\mathbf{e}}),
\end{aligned}
\end{equation}
where ${\mathbf{e}} = \frac{1}{\mathbf{T}_z}{\mathbf{t}}$.   
Given $\frac{h}{Z} = \gamma$ and $\frac{\mathbf{H} {\mathbf{p}}'}{\mathbf{H}_3 {\mathbf{p}}'}={\mathbf{p}}^w$, Eqn.~\ref{eq:d12} could be finally converted to
\begin{equation}
{\mathbf{p}}= (1-\frac{h \mathbf{T}_z}{Z h_c}){\mathbf{p}}^w +\frac{h \mathbf{T}_z}{Z h_c} {\mathbf{e}},
\end{equation}
\begin{equation}
(1-\frac{h \mathbf{T}_z}{Z h_c}) {\mathbf{p}} = (1-\frac{h \mathbf{T}_z}{Z h_c}){\mathbf{p}}^w +\frac{h \mathbf{T}_z}{Z h_c} {\mathbf{e}} - \frac{h \mathbf{T}_z}{Z h_c} {\mathbf{p}},
\end{equation}
\begin{equation}
{\mathbf{p}} -{\mathbf{p}}^w = \frac{-\frac{h \mathbf{T}_z}{Z h_c}}{1-\frac{h \mathbf{T}_z}{Z h_c}} ({\mathbf{p}} - {\mathbf{e}}),
\end{equation}
\begin{equation}
{\mathbf{p}} -{\mathbf{p}}^w = \frac{-\gamma \frac{\mathbf{T}_z}{h_c}}{1-\gamma \frac{ \mathbf{T}_z}{h_c}} ({\mathbf{p}} - {\mathbf{e}}).
\end{equation}
Applying the definition in Eqn.~\ref{eq:def}, we can calculate $\mathbf{u}_{res}$ from $\gamma$ by
\begin{equation}\label{eq:residual_flow}
  \mathbf{u}_{res}=\frac{-\gamma \frac{\mathbf{T}_z}{h_c}}{1-\gamma \frac{ \mathbf{T}_z}{h_c}} ({\mathbf{p}} - {\mathbf{e}}).
\end{equation}

\begin{figure*}[t!]
    \centering
    \includegraphics[width=\textwidth]{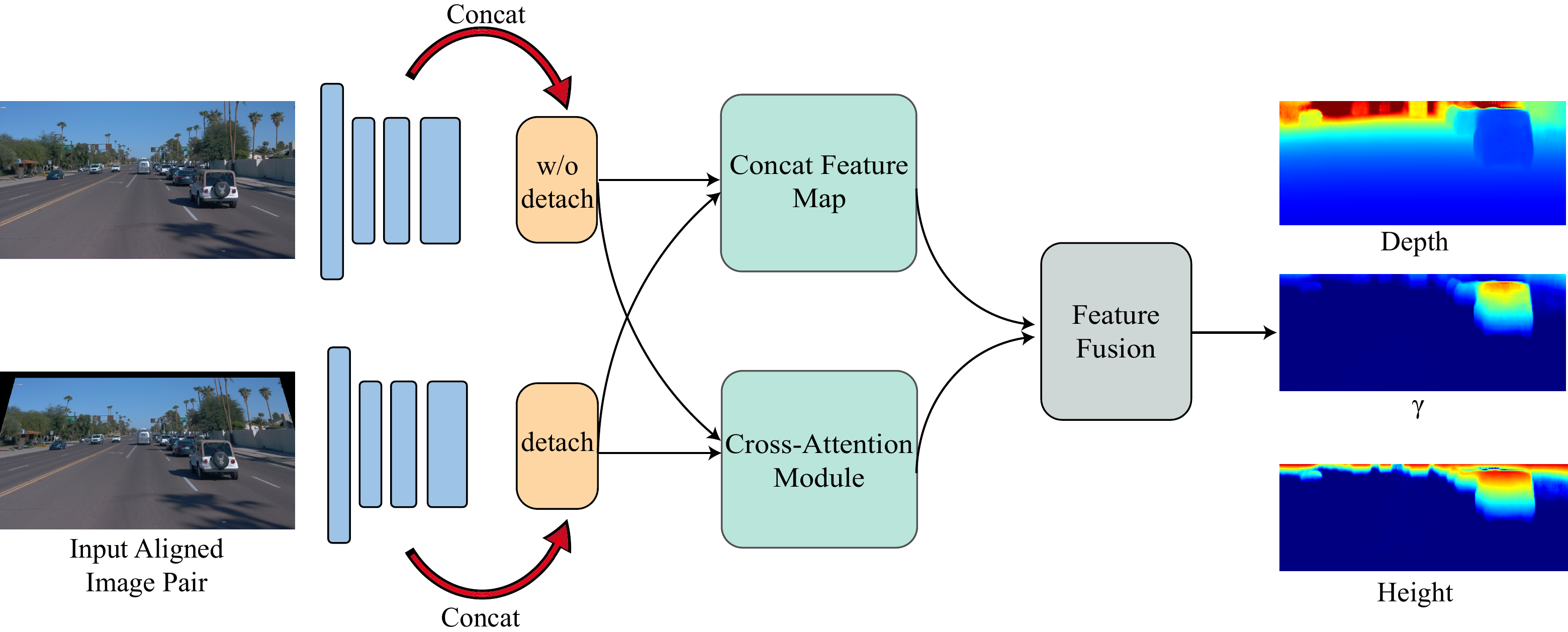}
    \caption{The proposed Road Planar Parallax Attention Neural Network.}
    \label{fig:arc}
\end{figure*}

From Eqn.~\ref{eq:residual_flow}, we know that the planar parallax can be easily gotten when the height of camera, the translation along z-axis, and $\gamma$ are available. In our framework, the $\gamma$ is estimated from the neural network, while others are from sensors or calibration. The benefit of the planar parallax is twofold. First, the homography perfectly depicts the underlying geometry of the road plane. After warping the source image $I_s$ via the road homography, pixels in road region would be aligned strictly with the target image $I_t$. In comparison, pixels in non-road regions would be affected by distortion of various degrees which is related to the height to the road plane. The distortion can provide vital cues for 3D reconstruction of the scene. Second, the homography can also remove the effect of rotation, which makes our method more robust to small baseline motion.

\paragraph{Road 3D Geometry Recovery}
Although our proposed method only uses 2D flow-based transformations to build the training loss, we can also perform 3D reconstruction from $\gamma$ (see Fig.~\ref{fig:geob} for details). 
Similar to Eqn.~\ref{eq:1}, we have
\begin{equation}\label{eq:1r}
   h = h_c - \vec{\mathbf{N}}^T\mathbf{P}.
\end{equation}
Supposing $\mathbf{K}$ is the camera's intrinsic matrix, $\mathbf{p}$ is the projection of $\mathbf{P}$ on image plane, whose 
homogeneous coordinate denoted by $\mathbf{p}=(x, y, 1)$, $\mathbf{P}$ could be calculated by an inverse projection given as  
\begin{equation}
\label{eq:inverse-pro}
    \mathbf{P} = Z\mathbf{K}^{-1}\mathbf{p}.
\end{equation}
Substituting Eqn.~\ref{eq:inverse-pro} into Eqn.~\ref{eq:1r} we could get
\begin{equation}\label{eq:2}
   h = h_c - \vec{\mathbf{N}}^T(\mathbf{K}^{-1}Z\mathbf{p}).
\end{equation}
Dividing both sides of Eqn.~\ref{eq:2} by $Z$ gives
\begin{equation}\label{eq:3}
   \frac{h}{Z} = \frac{h_c}{Z} - \vec{\mathbf{N}}^T(\mathbf{K}^{-1}\mathbf{p}).
\end{equation}
Defining $\gamma=\frac{h}{Z}$, Eqn.~\ref{eq:3} could be finally reorganized as 
\begin{equation}\label{eq:4}
   Z = \frac{h_c}{\gamma+\vec{\mathbf{N}}^T(\mathbf{K}^{-1}\mathbf{p})}.
\end{equation}
Note that the height of a pixel can be easily calculated by 
\begin{equation}\label{eq:hdconvert}
   h_p = \gamma Z.
\end{equation}
The above formulas theoretically proves that the planar parallax estimated by the deep neural network can be directly converted into height and depth of each pixel. In the experiment part, we will convert the planar parallax estimated by deep neural network into depth and height to verify our proposed method.

\begin{figure*}[ht]
\centering
\includegraphics[width=\linewidth]{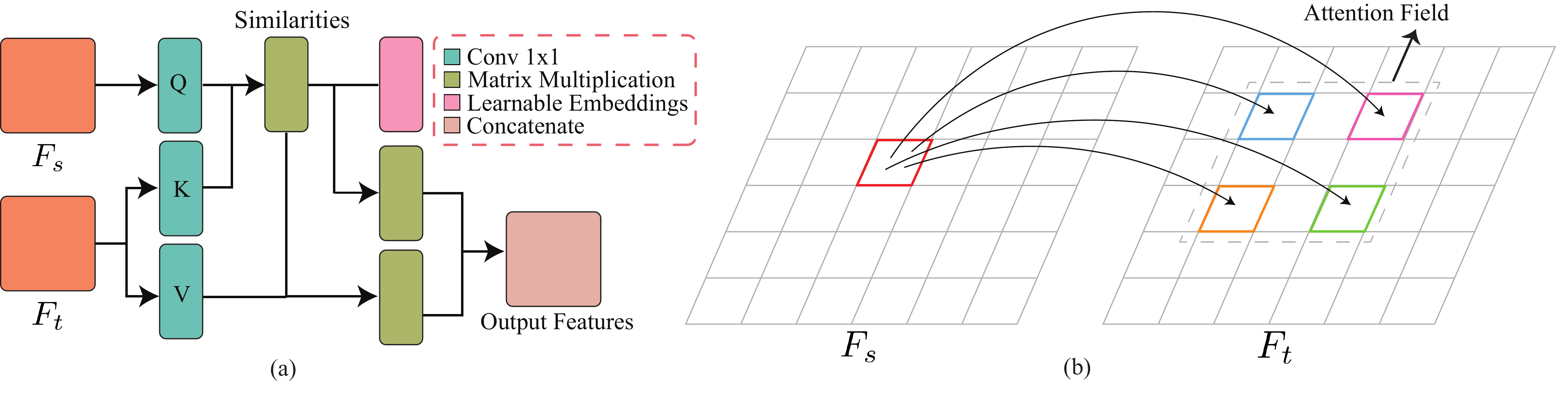}
\caption{(a) Proposed cross-attention module. (b) Illustration of cross-attention.}
\label{fig:ca}
\end{figure*}

\subsection{Network Architecture}
Planar parallax estimation requires perceiving the image displacements between two images. Based on this observation, in this section, we introduce RPANet, which consists of three modules: a CNN-based feature extraction layer, a cross-attention module, and an output layer. As shown in Figure.~\ref{fig:arc}, the input of RPANet includes two consecutive images, one of which is warped by road homography, and the output of RPANet is the $\mathbf{\gamma}$ map. The feature extraction module extracts features from both input images with shared weights. Similar to~\cite{chen2020exploring}, to avoid collapsing, we apply a stop-gradient operation on the feature extracted from warped image $I_s^w$. Then, the extracted features are fed into the cross-attention module, through which the key clues between the features can be modeled more effectively. Finally, a feature fusion module is applied to predict $\gamma$.

As illustrated in Fig.~\ref{fig:ca}(b), we adopt a cross-attention module. The cross-attention module tries to extract key clues for $\gamma$ estimation by performing neighborhood matching on two feature maps. We call the neighborhood area the attention field, which represents the field where one pixel of the feature map can be matched in the other feature map. Inspired by ~\cite{wang2020parallax}, we add a series of learnable parameters in order to make the network learn the implicit matching relationship efficiently.

As depicted in Fig.~\ref{fig:ca}(a), in our proposed cross-attention module, 1x1 convolution is firstly performed to extract $Q$, $K$, and $V$, which were described in \cite{vaswani2017attention}. The feature maps $F_s$ extracted from the source image are fed to a $1\times1$ convolution to produce $Q$. $K$ along with $V$ are extracted from $F_t$ which are feature maps extracted from the target image through another two $1\times1$ convolution modules. Then, partial matrix multiplication is conducted between $Q$ and $K$ where one pixel in $Q$ has a set of corresponding pixels in a fixed attention field of $K$.
A set of learnable parameters is utilized to find the matching relationship between the pixels from $F_s$ and $F_t$. Those parameters would be multiplied to the similarities generated from the result of partial matrix multiplication between $Q$ and $K$.
A similar process of partial matrix multiplication would be conducted between the similarities and $V$. 
For every pixel of output,
\begin{equation}
    y_o = \sum_{p \in A_{k \times k}} softmax(q_o^T*k_p)(v_p+r_{o,p}),
\end{equation}
where $y_o$ is one pixel of the output $O$, and $A_{k \times k}$ refers to the attention field.
$q_i$, $k_i$, and $v_i$ are vectors of a specific pixel from $Q$, $K$, and $V$ respectively, their shapes depend on the dimensions on channel. The $r_{a,b}$ is the learnable parameters represent a flow from $a$ to $b$ in two feature maps respectively. The partial matrix multiplication method is implemented with einsum and can be found in our implementation for details. Different from previous works \cite{li2020revisiting, wang2019learning} which build attention module based on the epipolar constraint, our cross-attention module is based on the local geometry of planar parallax. 

By applying a dilated attention on the 1/2 downsampled feature maps and $19\times19$ attention field, we avoid setting the attention field dense and global areas to reduce the number of parameters, which is similar to the dilated convolution \cite{yu2016multi}.

\subsection{Loss Function}
As the output $\gamma$ can be used to reconstruct a warped target image $I_t'$ from $I_s$, the widely used photometric loss can naturally be applied as supervisory signals. Besides, we use sparse ground truth $\gamma^*$ to build a sparse loss. Considering the photometric loss is not informative when applied on low-texture or homogeneous regions, we introduce additional smoothness loss to regularize our output. The total loss is given by:
 \begin{equation}
     E_{total} = \lambda_s E_s + \lambda_p E_p + \lambda_{sm} E_{sm},
 \end{equation}
where $E_s$ is sparse loss for $\gamma$ map, $E_p$ is photometric loss function, $E_{sm}$ is smoothness loss function. $\lambda_s$, $\lambda_p$ and $\lambda_{sm}$ are the loss weights on the respective loss term.

\paragraph{Photometric Loss Function}
After obtaining $\mathbf{\gamma}$ map, we can calculate the correspondence of each pixel between the source image $I_s$ and the target image $I_t$. Given the $\mathbf{u}_{res}$ of a pixel $\mathbf{p}_s$ in $I_s$,  we can get its corresponding pixel in $I_t$ as 
\begin{equation}
\begin{aligned}
    \mathbf{p}_t' &= \mathbf{u}_{\mathbf{p}} + \mathbf{H}_{s\to t} * \mathbf{p}_{s}\\
                 &=  \mathbf{u}_{\mathbf{p}} + \mathbf{p}_{s}^{w}.
\end{aligned}
\end{equation}
Based on the above equation, frame $I_{t'}$ can be reconstructed as 
\begin{equation}
    I_{t'}[\mathbf{p}_{t}] = I_{s} \langle \mathbf{p}_s	\rangle ,
\end{equation}
where $I_{t'}[\mathbf{p}_t]$ are pixel intensities at position $\mathbf{p}_t$, and $\langle \rangle$ is a bilinear sampling operator. 
Accordingly, the photometric loss function can be constructed to measure the difference between $I_t$ and  $I_t^{'}$. We use the robust photometric error combining SSIM~\cite{wang2004image} and L1 norm between two images which is given by
\begin{equation} \label{photo-loss-network}
    E_p(I_t, I_{t'}) = \alpha \cfrac{1-SSIM(I_t, I_{t'})}{2} + (1 - \alpha) ||I_t - I_{t'}||, 
\end{equation}
where $\alpha$ is a hyper-parameter.

\paragraph{Sparse Loss Function}
Since the  ground truth $\gamma^{*}$ can be generated from sparse LiDAR points, we directly use it to train the network. The sparse loss can be defined as
\begin{equation}
   E_s= \sum_{\mathbf{p}\in\Omega^l}|\gamma_{\mathbf{p}} - \gamma_{\mathbf{p}}^{*}|,  
\end{equation}
where $\gamma_{\mathbf{p}}$ and $\gamma_{\mathbf{p}}^{*}$ are the predicted and ground truth $\gamma$ value for pixel $\mathbf{p}$ respectively. $\Omega$ is the union of pixels that have ground truth.

\begin{figure*}[t!]
    \centering
    \includegraphics[width=\textwidth]{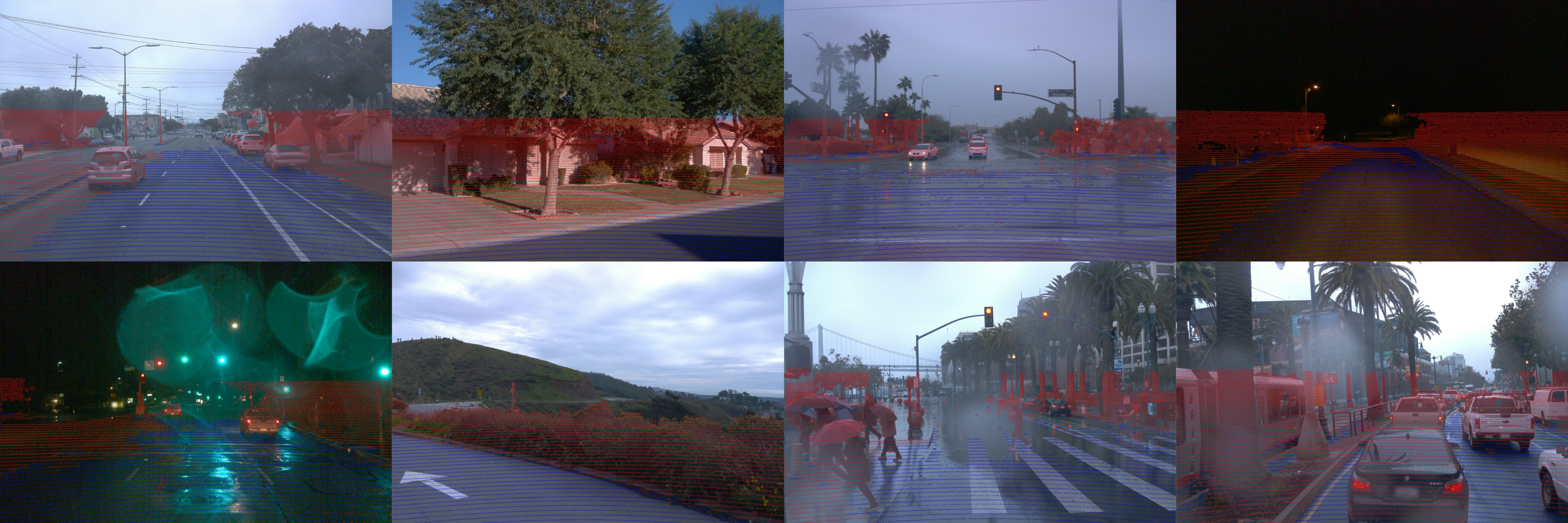}
    \caption{The Waymo Open Dataset used to construct RP2-Waymo dataset. The points in the figures are the LiDAR used for constructing the dataset. The blue points are the the points labeled with road. The road plane is extracted based on these points. Among all of the images, despite time, weather, scene vary, the road plane can be considered as a reference for planar parallax estimation.}
    \label{fig:datasets}
\end{figure*}

\paragraph{Smoothness Loss Function}
Edge-aware smoothness loss~\cite{godard2017unsupervised, bian2019unsupervised} is widely used by existing methods to enhance the depth's local consistency. Different from these methods, we apply the second order smoothness constraint~\cite{meister2018unflow, jonschkowski2020matters} on the residual flow. The smoothness loss function is defined as 
\begin{equation}
E_{sm} = \sum_{\mathbf{d}} \sum_{\mathbf{p}}\left(\left|\nabla^{\mathbf{d}} \mathbf{u}_{res}(\mathbf{p})\right|^{2} e^{-\beta\left|\nabla^{\mathbf{d}} I_{t}(\mathbf{p})\right|}\right),
\end{equation}
where $\nabla^{\mathbf{d}}$ stands for gradient calculated along the direction $\mathbf{d}$, $\beta$ is the weight for the gradient of image $I_t$ and $e$ is the natural base. In our method, we calculate the gradient of $\mathbf{u}_{res}$and $I_t$ in both horizontal and vertical direction. By leveraging the smoothness loss, the collinearity of neighboring flows is improved, and hence the final depth and height are regularized.

\section{Datasets}\label{sec:datasets}
Since there is no existing datasets dedicated to the planar parallax estimation task, we build a dataset named RP2-Waymo by carefully selecting data from the Waymo Open Dataset~\cite{sun2020scalability} and calculating the homography matrix. The RP2-Waymo dataset contains 13,030 training samples and 1,287 validation samples, which is challenging as it contains various scenes such as city, highway, suburb, and different weather conditions. 
To ensure fairness, we sample data uniformly from different sequences. The ground plane is extracted from the point cloud via robust algorithms such as RANSAC~\cite{fischler1981random}. Combined with odometry provided by the Waymo Open Dataset, the homography matrix needed by RPANet can be easily calculated.

\noindent
\textbf{Training Set.}
In the training set, we take full advantage of the LiDAR points to calculate the homography matrix, the road plane, and the $\gamma$ numbers of the pixels that are available. Each sample consists of two consecutive images, one is aligned by homography matrix, which we call the source image, and the other is called target image.

\noindent
\textbf{Validation Set.}
In the validation set, we create two modes to evaluate the proposed method thoroughly in order to measure the performance of different use cases in the real world. 

a). The road plane is available (PA). In this setting, we use LiDAR to construct not only the ground truth of $\gamma$, depth, and height, but also the road plane and homography matrix. We evaluate our methods on this setting because this setting can help us evaluate the deep neural network excluding the errors caused by the homography matrix and road plane calibration as much as possible. In the real world, the homography matrix and road plane can be calibrated with sensors inside the car. We leave the analysis of errors caused by these sensors for future work, since the sensors are not available in the Waymo Open Dataset now.

b). The road plane is not avaliable (PNA). In this setting, we only use LiDAR to construct the ground truth of $\gamma$, depth, and height for evaluation. In other words, the 3D reconstruction does not require LiDAR at all. In order to generate the road plane and homography matrix, we apply a homographynet with ResNet-18 [54] backbone and two head each contains 3 convolution layers. The homographynet takes raw images as input and output the road plane and pose, then the homography can be computed from them. When training the homographynet, we apply cosin ssimilarity loss for ground norm and photometric loss for homography matrix. We are surprised to find that a very simple network can get very effective results, although it will cause acceptable errors. More details can be found in Sec.\ref{sec:exp}.

\begin{table*}
\normalsize
\caption{\textbf{The mean absolute error of height and depth.} Depth and height are obtained from $\gamma$ according to Eqn.~\ref{eq:depth}. ``\textbf{Geometry}'' refers to using the traditional geometry based method. ``\textbf{Depth-Baseline-R18}'' represents a depth estimation baseline with ResNet-18 backbone (height results are calculated by Eqn.~\ref{eq:hdconvert} with road plane ground truth). ``\textbf{GammaNet-R18}'' is a simple baseline that has the same network structure but predict gamma supervised by ground truth. ``\textbf{w/o U}'' refers to our RPANet without the unsupervised loss (Eqn.~\ref{photo-loss-network}). For the validation set, ``\textbf{PA}'' refers to that road plane is available and ``\textbf{PNA}'' indicates that road plane is not available. The best results are shown in bold.}
\begin{center}
\begin{tabular}{l|c|cccc|ccc}
\toprule
\multirow{2}{*}{Method}& \multirow{2}{*}{Validation Set}& \multicolumn{4}{c}{Absolute Height Error(m)} & \multicolumn{3}{c}{Absolute Depth Error(m)} \\
&& $h<0.1 m$ & $h<0.3m$ & $h<0.5 m$ & $h<1 m$       & $d<30m$ & $d<50m$ & $d<80m$ \\
\hline
\textbf{Geometry}   &\textbf{PA}   &0.290 &0.420 &0.513 &0.603         &2.82 &10.80  &14.93\\
\hline
\textbf{Depth-Baseline-R18}   &\textbf{PA}   &0.057 &0.066 &0.069 &0.077         &0.545 &0.947   &1.352\\

\textbf{GammaNet-R18}    &\textbf{PA}  &0.021 &0.034 &0.041 &0.056          & 0.388   &0.775    &1.200  \\

\textbf{RPANet w/o ``U''}  &\textbf{PA}   &0.022 &0.034 &0.041 &0.053           & 0.358   & 0.703   & 1.162 \\

\textbf{RPANet}  &\textbf{PA}  &\textbf{0.019} &\textbf{0.031} &\textbf{0.038} &\textbf{0.051}           &\textbf{0.337}   &\textbf{0.702}   &\textbf{1.140} \\
\hline
\textbf{RPANet} &\textbf{PNA} &0.023 &0.036  &0.043 &0.057 &0.362 &0.772 &1.252\\
\bottomrule
\end{tabular}
\label{tab:ablation}
\end{center}
\end{table*}

\noindent
\textbf{Metrics.}
We use the Mean Absolute Error (MAE) to evaluate the proposed methods on height and depth. 
\begin{equation}\label{eq:mae}
    MAE = \frac{\sum_i^n|\hat{y}_i - y_i|}{n}.
\end{equation}
In Eqn.~\ref{eq:mae}, only pixels with ground truth are calculated in. The height and depth are also evaluated under different depth and height intervals to report the range where errors happen. Following~\cite{godard2019digging}, we also apply the widely used metrics in depth estimation evaluation to compare with methods for comparison. It is worth noting that height and $\gamma$ cannot be measured by relative errors, because the ground truth may have zero or negative values. For this reason, we do not apply relative metrics (e.g. the absolute relative error) on $\gamma$ and height evaluation.

\section{Experiments} \label{sec:exp}
In this section, we evaluate the proposed method by comparing it with the depth estimation methods on our proposed datasets. After that, we give an analysis of the performance of these methods.

\subsection{Implementation Details} \label{sec:impl}
Our framework is implemented using Pytorch~\cite{paszke2019pytorch}. We adopt Adam optimizer~\cite{kingma2015adam} with $\beta_1$ = 0.9 and $\beta_2$ = 0.999 to update the parameters of RPANet. All the experiments are performed on a stand-alone server with 4 NVIDIA TITAN Xp GPUs. With the setting of corresponding default hyper-parameters described above, each training step costs about 0.7 seconds.
In the default settings, we use 3 GPUs with a batch size of 6 and each GPU needs about 8GB RAM though it has about 12GB RAM. Our model is trained for 20 epochs on the training set with the learning rate reduced by a factor of 10 after the tenth epoch where the initial learning rate is 0.0001. All hyper-parameters are tuned based on a 1000-size development set, and we make no further adjustments except for the number of epochs for training on the complete training set. The input images are first cropped (from $1920 \times 1280$ to $1920 \times 1024$) and then resized to  $960 \times 512$ using the Lanczos interpolation algorithm. We adopt a ResNet-18 backbone following~\cite{godard2019digging}.

\noindent
\textbf{Baselines.}
Since our method mainly focuses on the planar parallax geometry and the cross-attention module for finding the relationship between consecutive images, it can be adapted with different feature extractors. We adopt a ResNet-18 backbone following~\cite{godard2019digging} to build RPANet. As there lacks planar parallax methods to be compared with, to verify the effectiveness of RPANet, we also build a baseline depth estimation with ResNet-18 backbone (denoted as ``\textbf{Depth-Baseline-R18}'') for fair comparison. Note that the Depth-Baseline-R18 is with full supervision rather than self supervision in~\cite{godard2019digging}. To compare our method with recent transformer-based depth estimation method, we also build a ``\textbf{Depth-Baseline-DPT}'' beyond the DPT-Hybrid~\cite{ranftl2021vision} feature extractor. To compare our method with a naive baseline that predicts gamma map with supervision, we use a ``\textbf{GammaNet-R18}'' with the same network structure as ``\textbf{Depth-Baseline-R18}'' but predicts gamma by supervision. We also provide a geometry-only method ``\textbf{Geometry}'' to compare RPANet with traditional methods. Specifically, we estimate the optical flow with OpenCV and calculate the depth and height with planar parallax geometry as described in Sec.~\ref{sec:planar_parallax}, which is similar to the practice in the traditional planar parallax methods~\cite{kumar1994direct, sawhney19943d}.

\begin{table*}
\normalsize
\caption{Ablation Study on the proposed RPANet. RPANet is our proposed method. "\textbf{w/o RE}" means that without the relative embedding in our proposed method. "\textbf{w/o DE}" means that without the detachment after the feature of the warped image in the network. All of the results in this table are gotten under \textbf{PNA}.}
\begin{center}
\begin{tabular}{l|c|cccc|ccc}
\toprule
\multirow{2}{*}{Method}& \multicolumn{4}{c}{Absolute Height Error(m)} & \multicolumn{3}{c}{Absolute Depth Error(m)} \\
& $h<0.1 m$ & $h<0.3m$ & $h<0.5 m$ & $h<1 m$       & $d<30m$ & $d<50m$ & $d<80m$ \\
\hline
\textbf{RPANet}  &0.019 &0.031 &0.038 &0.051          &0.337  &0.702   &1.140 \\
\hline
\textbf{RPANet w/o RE} &0.039 &0.050 &0.056 &0.067 &0.475 &0.856 &1.297  \\
\textbf{RPANet w/o DE} &0.030 &0.041 &0.047 &0.059 &0.405 &0.788 &1.228  \\
\bottomrule
\end{tabular}
\label{tab:ablation2}
\end{center}
\end{table*}

\subsection{Quantitative Results}
\begin{table*}
\normalsize
\caption{The mean absolute error of height and depth for \textbf{RPANet} and \textbf{Depth-Baseline-R18} in different depth and height intervals. $h_0/d_0$ means that in the specific height and depth range, $h_0\:m$ and $d_0\:m$ are the mean absolute error of height and depth respectively.}
\begin{center}
\begin{tabular}{c|c|cccc}
  \toprule
  \diagbox{Depth}{Height}&Method&$h<0.1m$ & $h<0.3m$ & $h<0.5m$ & $h<1m$\\
  \midrule
  \multirow{2}*{$d<30m$}& \textbf{RPANet} & $0.014/0.13$ & $0.020/0.19$ & $0.022/0.22$ & $0.029/0.34$ \\
  ~ & \textbf{Depth-Baseline-R18} &$0.049/0.39$ & $0.053/0.45$ & $0.054/0.47$ & $0.058/0.55$ \\
  \hline
  \multirow{2}*{$d<50m$}& \textbf{RPANet} & $0.017/0.20$ & $0.026/0.36$ & $0.031/0.45$ & $0.041/0.70$ \\
  ~ & \textbf{Depth-Baseline-R18} & $0.055/0.55$ & $0.061/0.69$ & $0.063/0.76$ & $0.069/0.95$ \\
  \hline
  \multirow{2}*{$d<80m$}& \textbf{RPANet} & $0.019/0.27$ & $0.031/0.52$ & $0.038/0.69$ & $0.051/1.14$\\
  ~ & \textbf{Depth-Baseline-R18} &$0.057/0.64$ & $0.066/0.88$ & $0.069/1.02$ & $0.077/1.35$\\  
  \bottomrule
\end{tabular}
\label{tab:Height1_2}
\end{center}
\end{table*}

\begin{table*}[ht!]
\normalsize
\caption{Comparison results of our method and the representative depth estimation method. The definitions of metrics are same as \cite{godard2019digging}. ``\textbf{RPANet + DPT}'' refers to a DPT-Hybrid~\cite{ranftl2021vision} backbone adapted into our RPANet. The metrics with \textcolor{orange}{orange} background means ``\textbf{lower} is better''. The metrics with \textcolor{blue}{blue} background means ``\textbf{higher} is better''}
\begin{center}
\begin{tabular}{|l||c|c|c|c|c|c|c|}
\hline
method & \cellcolor{orange}Abs Rel & \cellcolor{orange}Sq Rel & \cellcolor{orange}RMSE & \cellcolor{orange}RMSE log & \cellcolor{blue}{\scriptsize $\sigma < 1.25$} & \cellcolor{blue}{\scriptsize $\sigma < 1.25^2$} & \cellcolor{blue} {\scriptsize $\sigma < 1.25^3$} \\
\hline
\textbf{Depth-Baseline-R18}& 0.0474 & 0.3302 & 3.505 & 0.0876 & 0.970 & 0.992 & 0.997\\
\hline
\textbf{RPANet}& 0.0378 & 0.4491 & 3.934 &  0.0896 & 0.964 & 0.989 & 0.996\\
\hline
\textbf{Depth-Baseline-DPT}~\cite{ranftl2021vision}& 0.0247 & 0.2573 & 2.110 & 0.0732 & 0.982 & 0.997 & 0.999\\
\hline
\textbf{RPANet + DPT}& 0.0201 & 0.2660 & 2.209 &  0.0755 & 0.986 & 0.997 & 0.999\\
\hline
\end{tabular}

\label{tab:comparision}
\end{center}
\end{table*}

\noindent
\textbf{Ablation Study.} 
As shown in Table.\ref{tab:ablation} and Table.\ref{tab:ablation2}, we conduct ablation study of our method on the RP2-Waymo dataset. The RPANet is trained on the training set in supervised or semi-supervised manner, and evaluated on the two validation sets including \textbf{PA} (road plane is available through LiDAR) and \textbf{PNA} (road plane is not available, which means LiDAR is not used during inference). As expected, our RPANet containing the proposed cross-attention module as well as trained with all the above loss functions, achieves the best results. With the depth in the range of $0-80$ meters, our proposed RPANet has achieved a mean absolute error of \textbf{1.140} which is the best results among all of the methods. Note that even without road plane (the PNA setting), RPANet still outperform \textbf{Depth-Baseline-R18} with a large margin.

The results of different networks are reported in in Table.\ref{tab:ablation}. We can see that the traditional geometry based method is almost failed in our dataset and the accuracy of ``\textbf{Depth-Baseline-R18}" is significantly lower than the others predicting $\gamma$ especially in height estimation, which validates the effectiveness of our proposed method. We can also notice that the results of network ``\textbf{GammaNet-R18}'' is comparative to that of network ``\textbf{RPANet w/o U}'' with respect to depth metric while  worse in the height metric. The results indicate that the cross-attention module may provide more useful information for depth evaluation. Comparing the results of ``\textbf{RPANet w/o U}'' and ``\textbf{RPANet}'' in Table.\ref{tab:ablation}, we can conclude that the photometric loss reduces the MAE in all intervals regardless of height and depth. This is because that the photometric loss helps RPANet learn more accurate correspondences  and supply complementary supervision for areas lack of ground truth.

In Table.\ref{tab:ablation2}, we do the ablation studies on some strategies in our proposed RPANet. The results show that both the relative embedding in the cross-attention module and the detachment after the feature of the homography warped image are very useful for the network. The former is because that the proposed network highly relies on finding the displacement between homography aligned images to estimate the $\gamma$ map. The latter is to prevent the whole network collapses.

As shown in Table.\ref{tab:Height1_2}, we also report the detailed mean absolute error of height and depth in different intervals of depth and height for providing more information of our final setting. From Table.~\ref{tab:Height1_2}, we can see that the results of both networks become poor for objects that are farther and higher. This is because the ground truth in the distance is more sparse, and the distance target is too small to estimate the matching relationship.

\noindent
\textbf{Comparison between PA and PNA.}
During training, we can use LiDAR to obtain the accuracy road plane and homography matrix, but during inference we may only have image sequences. To validate the utility of our method, we further compare the results of \textbf{PA} and \textbf{PNA} mentioned in Sec.~\ref{sec:datasets}. Comparing the results of height in \textbf{PNA} and \textbf{PA} of our method, \textbf{PNA} is worse than \textbf{PA} of all setting, but the gap is within an acceptable range. When comparing the results of depth, the \textbf{PNA} of ``\textbf{RPANet}'' is better than the result of ``\textbf{Depth-Baseline-R18}'', this indicates that unsupervised loss and attention are more efficient for depth. Further more, even the plane and homography information supplied from a simple network, the result of \textbf{PNA} is better than ``\textbf{Depth-Baseline-R18}'' with a large margin.

\begin{table*}[ht!]
\normalsize
\caption{Comparison of different scenes in absolute height error and absolute depth error.}
\begin{center}
\begin{tabular}{c|cccc|ccc}
\toprule
\multirow{2}{*}{Scene}& \multicolumn{4}{c}{Absolute Height Error(m)} & \multicolumn{3}{c}{Absolute Depth Error(m)} \\
            & $h<0.1 m$ & $h<0.3m$ & $h<0.5 m$ & $h<1 m$       & $d<30m$ & $d<50m$ & $d<80m$ \\
\hline
city      &0.022 &0.032 &0.038 &0.049         &0.361 &0.700   &0.993\\

suburb      &0.019 &0.030 &0.035 &0.043          & 0.224   &0.537    &1.000  \\

highway     &0.012 &0.018 &0.024 &0.041           & 0.299   & 0.571   & 1.005 \\

night   &0.027 &0.056 &0.074 &0.122       &1.003   &1.873   &2.469 \\
\bottomrule
\end{tabular}
\label{tab:scenes}
\end{center}
\end{table*}

\noindent
\textbf{Comparison with Depth Estimation.} To fully evaluate the effectiveness of our method, comparative experiments with the depth estimation network are conducted and the results are reported in Table.~\ref{tab:comparision}. The ground truth of depth is generated from LiDAR points and used as sparse supervision. As reported in Table.~\ref{tab:comparision}, the proposed RPANet outperforms the depth estimation networks both with CNN (ResNet-18~\cite{he2016deep}) and transformer (DPT-Hybrid~\cite{ranftl2021vision}) backbone in absolute relative error but lags behind on square-based errors. We speculate that this may be due to the complexity of the scenes. In the complex scenes, our RPANet is affected by error of both plane estimation and gamma estimation at relatively far pixels. Considering the depth map is got from Eqn.~\ref{eq:4}, only a small error in the gamma and road plane will lead to a large error in depth map. For the far pixels, we would like to leave the more accurate depth estimation as future work. For example, one possible solution is combining the advantages of depth estimation and planar parallax and have better performance at both shorter distances and longer distances. But we still want to emphasize that depth estimation of pixels with a relatively close distance (e.g., less than 30m) may be more useful for driving scenes, and our RPANet has a pretty good performance in those pixels. This is due to the benefit brought by the geometric constraint of planar parallax, which makes the neural network learn to predict much easier.

\begin{table}
\normalsize
\caption{The number of each scene in the validation set.}
\begin{tabular}{c|cccc}
  \toprule
  Scenes& city & suburb & highway & night\\
  \midrule
  Number of Samples & 249 & 718 & 197 & 123\\
  \bottomrule
\end{tabular}
\label{tab:scene_num}
\end{table}

\noindent
\textbf{Comparison of Different Scenes.}
To evaluate the performance of our method on different scenes, the validation set of RP2-Waymo dataset are clustered into 4 categories : \textbf{city}, \textbf{suburb}, \textbf{highway}, and \textbf{night}. The distribution of the validation set in the four scenarios and the original dataset is kept consistent without deliberate adjustments. The number of samples in each scene can be found in Table~\ref{tab:scene_num}. We test our proposed RPANet under all 4 scenes and the results are reported in Table.~\ref{tab:scenes}. It can be seen from the data that in the night scene, the performance of our proposed RPANet has degraded. this is possibly because that the image quality at night is poor and there are fewer clues to recover the three-dimensional information from the image.

\begin{figure}
\includegraphics[width=\linewidth]{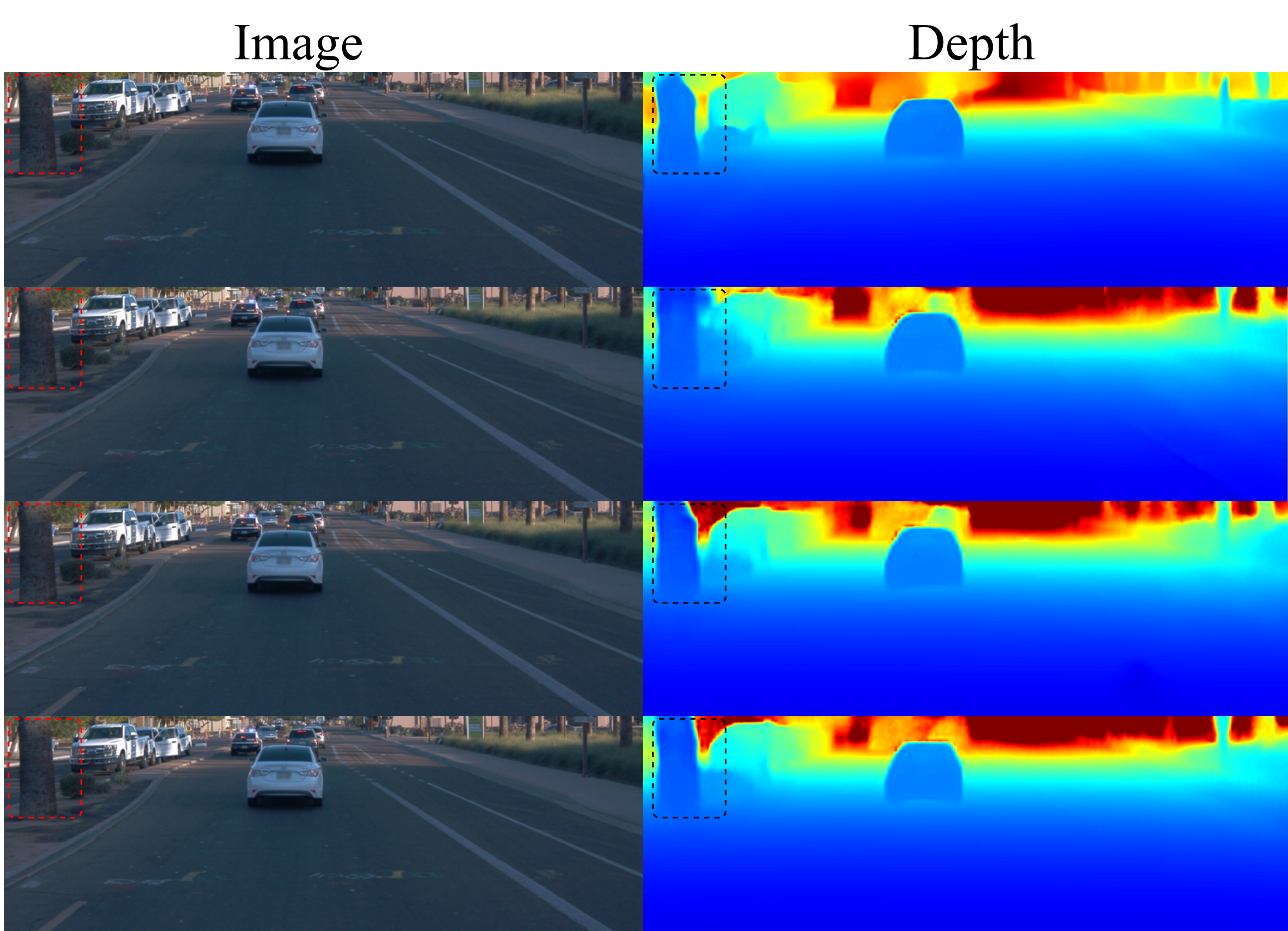}
\caption{Comparison on depth of different methods. Coloring from blue to red represents depth value from small to large. Rectangles depict regions for comparison. From top to bottom: (1).``\textbf{Depth-Baseline-R18}''; (2).``\textbf{GammaNet-R18}; (3).``\textbf{RPANet w/o U}''; (4).``\textbf{RPANet}''.}
\label{fig:illu2}
\end{figure}

\begin{figure*}
\centering
\includegraphics[width=\linewidth]{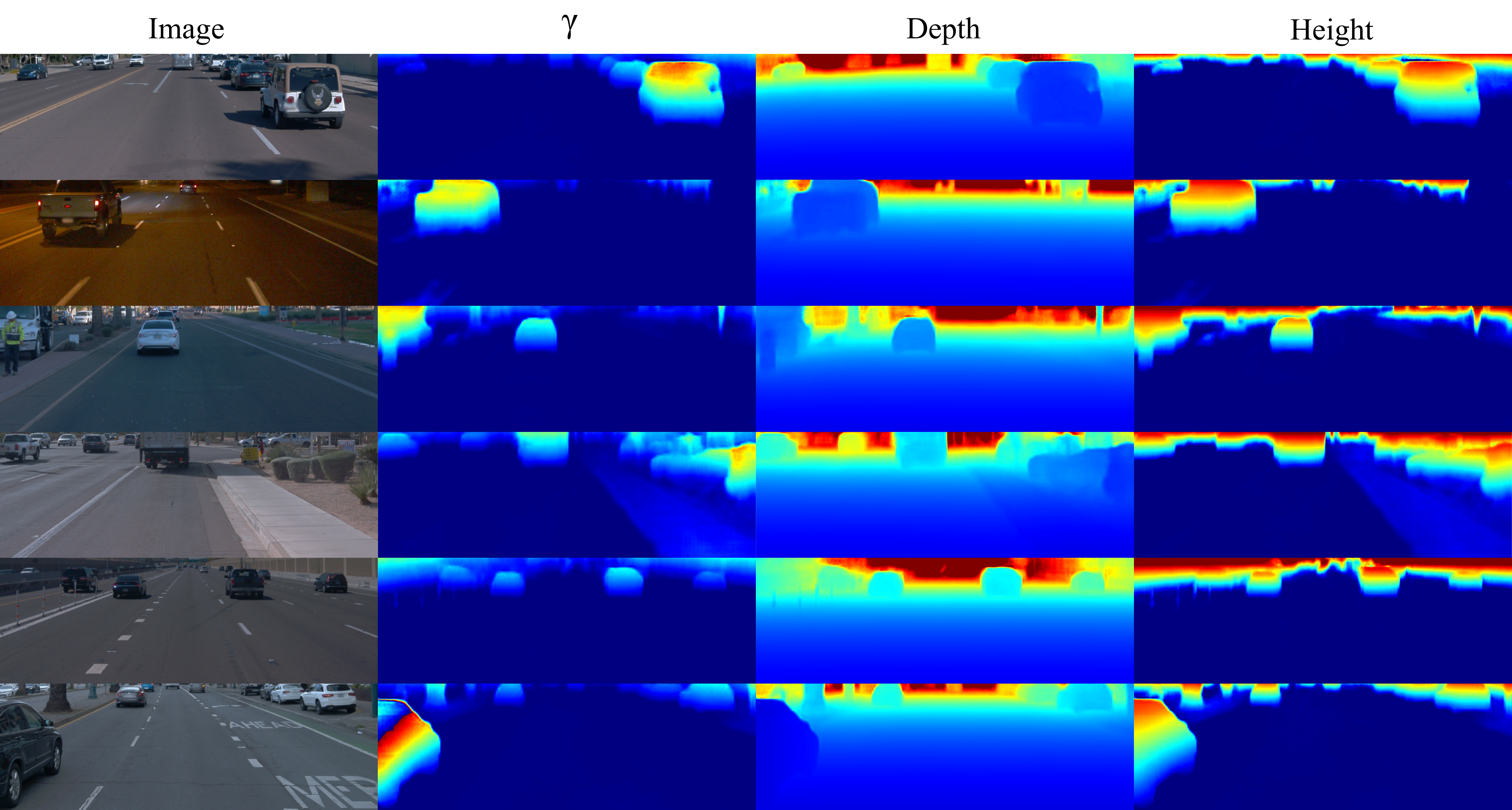}
\caption{The visualization results of $\gamma$, depth, and height of different inputs.} \label{fig:illu}
\end{figure*}

\begin{figure*}
    \subfloat[]{\includegraphics[width=.25\textwidth]{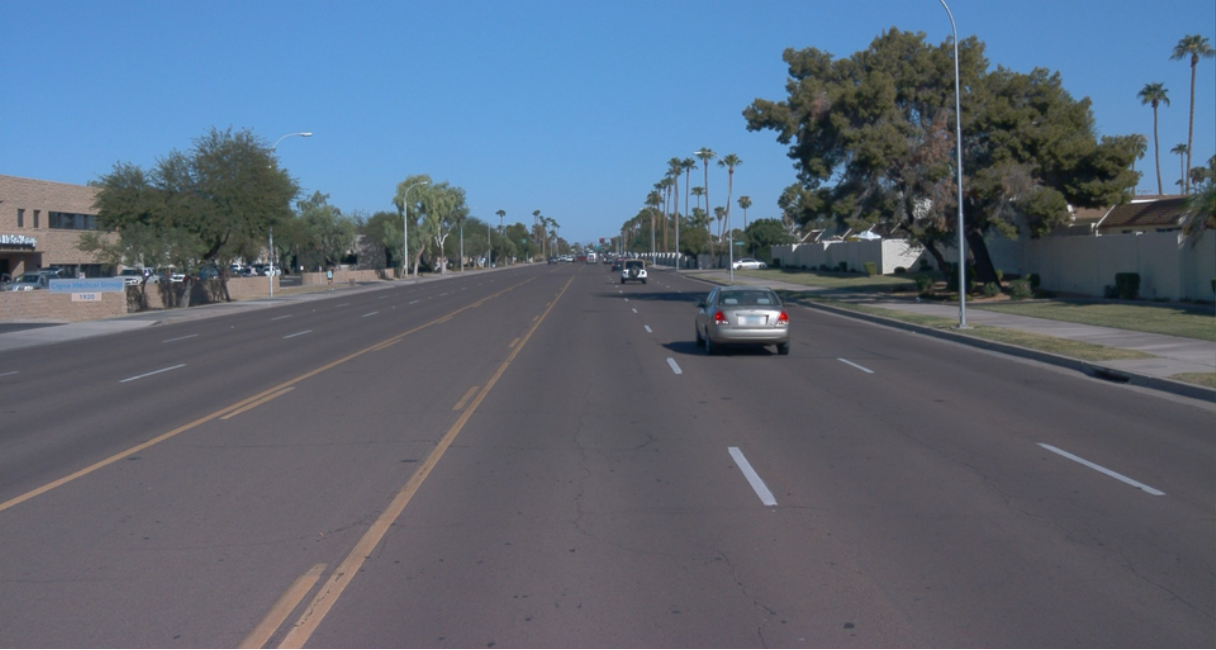}%
    \label{fig:illu3_1}}
    \hfil
    \subfloat[]{\includegraphics[width=.25\textwidth]{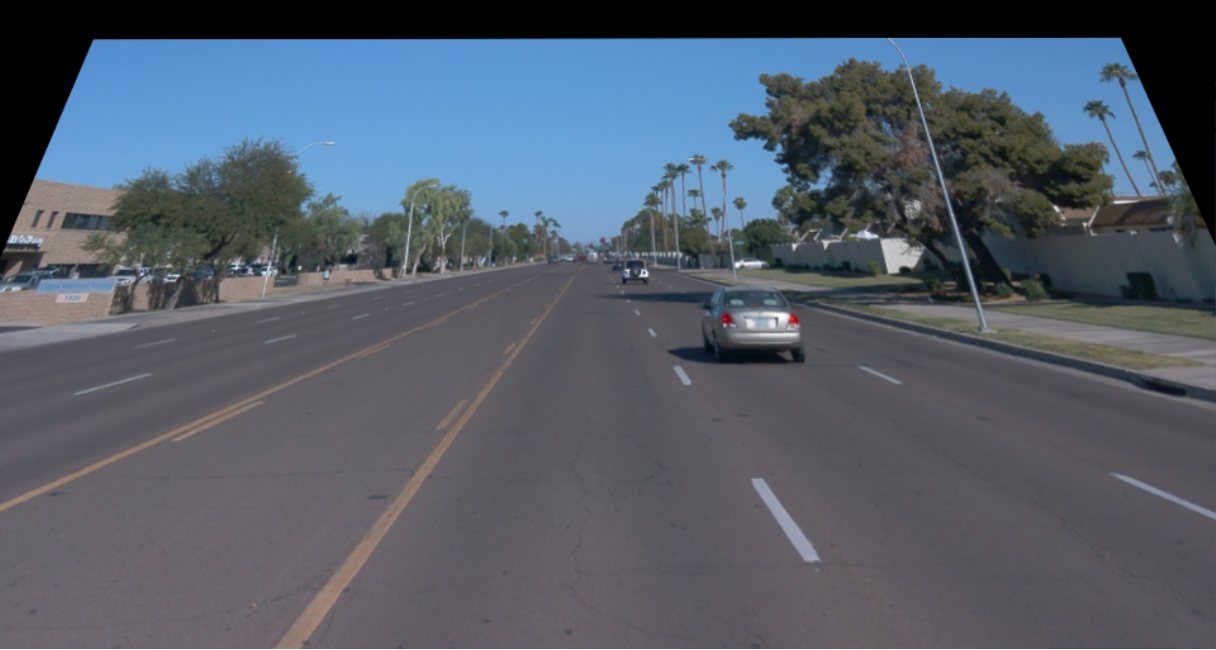}%
    \label{fig:illu3_2}}
    \hfil
    \subfloat[]{\includegraphics[width=.25\textwidth]{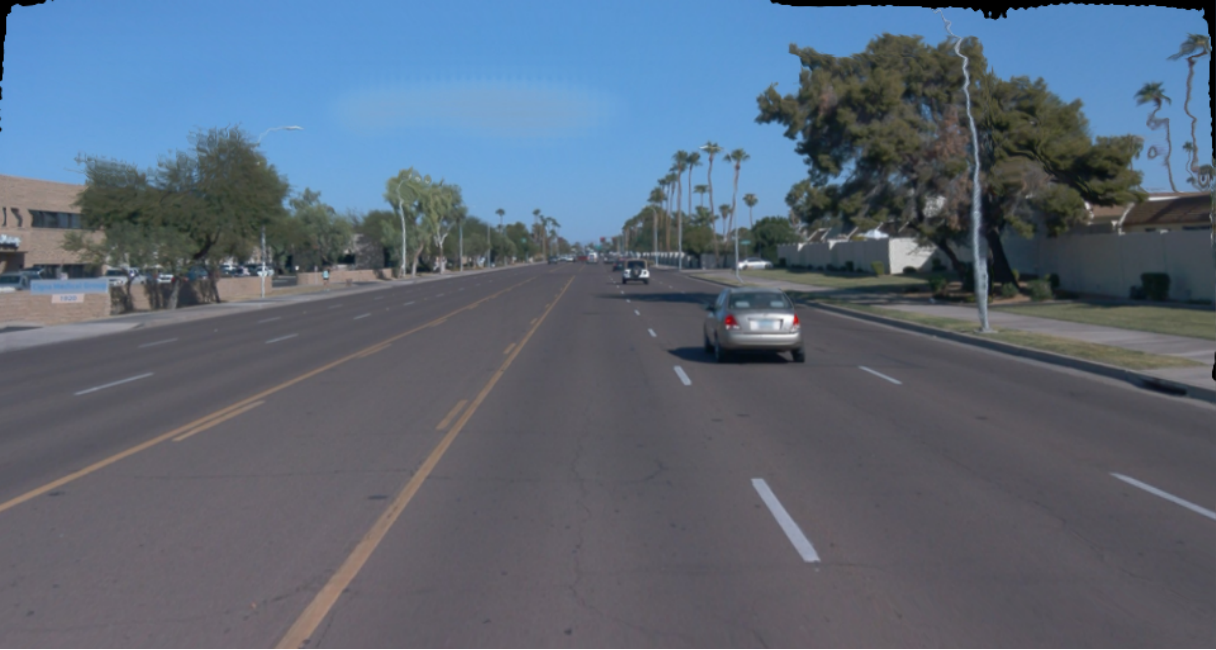}%
    \label{fig:illu3_3}}
        \hfil
    \subfloat[]{\includegraphics[width=.25\textwidth]{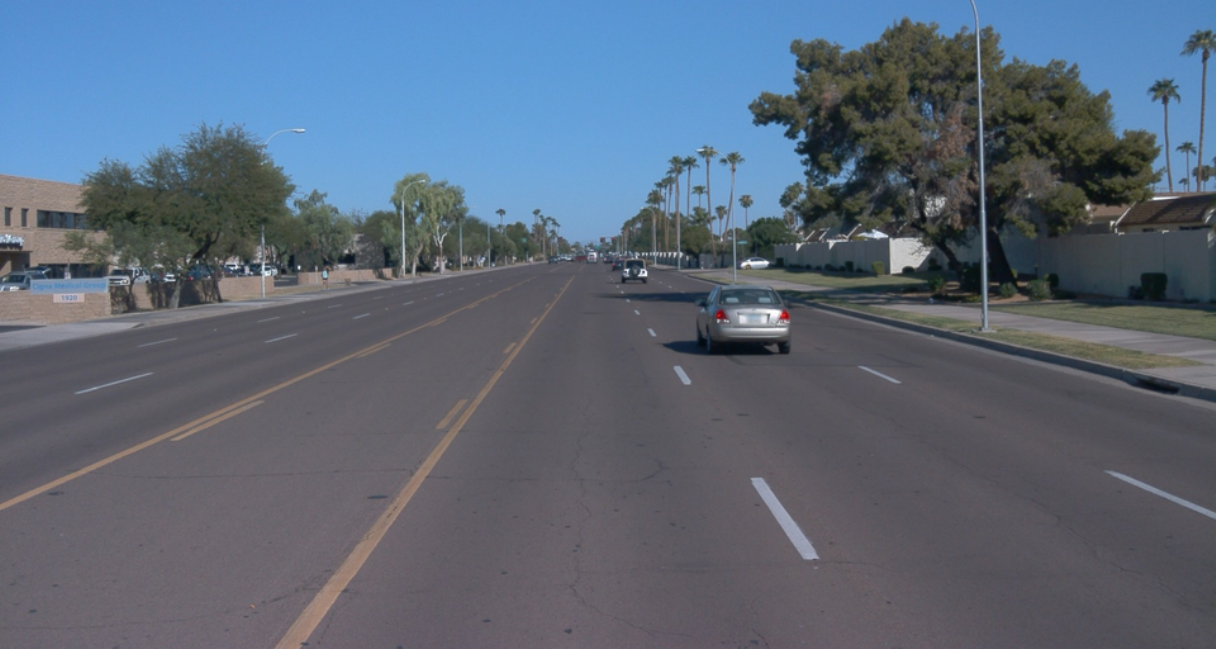}%
    \label{fig:illu3_4}}
  \caption{(a) Source image $I_s$. (b) Homography warped source image $I_s^w$. (c) Reconstructed image $I_t'$. (d) Target image $I_t$. (a) to (b) refers to the homography aligning process; (b) to (c) refers to the planar parallax reconstruction process; (c) to (d) illustrates the difference between reconstructed image and the target image.}
  \label{fig:illu3}
\end{figure*}

\subsection{Qualitative Results}
The output of RPANet are represented in Fig.~\ref{fig:illu}. The $\gamma$ is the direct result of the RPANet, while the depth and height are converted from $\gamma$ according to Eqn.~\ref{eq:depth}. Colors from blue to red represents values from small to large. 
It can be seen from the $\gamma$ map that the $\gamma$ value of the nearby vehicle is relatively large, while the distant vehicle is relatively small, which is consistent with the definition of $\gamma$. Compared with depth information, $\gamma$ and height information can better distinguish obstacles above the road such as sidewalks, which to a certain extent is essential for autonomous or assisted driving. 
From the results, we can see that the drivable surface can be easily identified with the help of height and depth extracted from $\gamma$.

\noindent
\textbf{Comparison of Different Methods.}
As shown in Fig.~\ref{fig:illu2}, we compare the different methods with the visualization results.
The results show that RPANet could predict accurate depth, especially in discontinuous regions (see the boundaries of trees in rectangles). At the edge of the tree, the illumination of it is relatively low, making depth prediction difficult. Nonetheless, after applying the geometric constraint of planar parallax, the depth information can be well calculated leveraging $\gamma$.
We take the tree besides the road as an example. Although the depth estimation network directly outputs the depth and thus avoiding the error caused by the road surface calibration or magnified by Eqn.~\ref{eq:depth}, the depth estimation network does not identify the boundary of the tree successfully. 
"\textbf{GammaNet-R18}" does not distinguish between the tree and the road surface next to it, while RPANet distinguishes the tree from the pavement better. It shows that the cross-attention module and the photometric loss help the neural network to learn $\gamma$ more easily.

\noindent
\textbf{Image Reconstruction via Residual Flow.} As illustrated in Fig.~\ref{fig:illu3}, the road plane of source image and target image are aligned after homography warping, while other static areas can be further aligned by the residual flow warping. The visualization shows that the two warping steps build a bridge between the source image and the target image, due to that we can easily obtain the reconstructed image and leverage photometric loss to train our RPANet. In Fig.~\ref{fig:illu3}, it is easy to find whether the proposed framework has successfully reconstructed the 3D scene with planar parallax. From (a) to (b), the road is well aligned with homography matrix while things above the road are with a displacement. From (b) to (c), things above the road are corrected with planar parallax estimated by neural network. The difference between (c) and (d) indicates that errors still exist, especially for the pixels with a large height or depth in the image. The well-aligned pixels from (c) to (d) indicate that their 3D structures are well estimated.

\section{Conclusion}
In this paper, we propose a planar parallax estimation method, which combines neural networks and planar parallax geometry. The input of our method is aligned image pairs via road homography. The output $\gamma$ map is utilized to recover the 3D structure (depth and height). We also devise the cross-attention module to learn planar parallax more easily. Since no public dataset provides aligned images via road homography, we collect data from the Waymo Open Dataset and build the RP2-Waymo dataset. Comprehensive experiments conducted on the datasets valid the effectiveness of our method.

\bibliographystyle{IEEEtran}
\bibliography{main}

\begin{thebibliography}{10}
\providecommand{\url}[1]{#1}
\csname url@samestyle\endcsname
\providecommand{\newblock}{\relax}
\providecommand{\bibinfo}[2]{#2}
\providecommand{\BIBentrySTDinterwordspacing}{\spaceskip=0pt\relax}
\providecommand{\BIBentryALTinterwordstretchfactor}{4}
\providecommand{\BIBentryALTinterwordspacing}{\spaceskip=\fontdimen2\font plus
\BIBentryALTinterwordstretchfactor\fontdimen3\font minus
  \fontdimen4\font\relax}
\providecommand{\BIBforeignlanguage}[2]{{%
\expandafter\ifx\csname l@#1\endcsname\relax
\typeout{** WARNING: IEEEtran.bst: No hyphenation pattern has been}%
\typeout{** loaded for the language `#1'. Using the pattern for}%
\typeout{** the default language instead.}%
\else
\language=\csname l@#1\endcsname
\fi
#2}}
\providecommand{\BIBdecl}{\relax}
\BIBdecl

\bibitem{asai20083d}
T.~Asai, K.~Yamaguchi, Y.~Kojima, T.~Naito, and Y.~Ninomiya, ``{3D} line
  reconstruction of a road environment using an in-vehicle camera,'' in
  \emph{Proc. ISVC}, 2008, pp. 897--904.

\bibitem{chen2015fast}
D.~Chen and X.~He, ``Fast automatic three-dimensional road model reconstruction
  based on mobile laser scanning system,'' \emph{Optik}, vol. 126, no. 7-8, pp.
  725--730, 2015.

\bibitem{gao20183d}
H.~Gao, L.~Liu, Y.~Tian, and S.~Lu, ``3d reconstruction for road scene with
  obstacle detection feedback,'' \emph{IJPRAI}, p. 1855021, 2018.

\bibitem{badue2020self}
C.~Badue, R.~Guidolini, R.~V. Carneiro, P.~Azevedo, V.~B. Cardoso, A.~Forechi,
  L.~Jesus, R.~Berriel, T.~M. Paixao, F.~Mutz \emph{et~al.}, ``Self-driving
  cars: A survey,'' \emph{ESWA}, p. 113816, 2020.

\bibitem{schoettle2014survey}
B.~Schoettle and M.~Sivak, ``A survey of public opinion about autonomous and
  self-driving vehicles in the us, the uk, and australia,'' University of
  Michigan, Ann Arbor, Transportation Research Institute, Tech. Rep., 2014.

\bibitem{lu2020autonomous}
B.~Lu, B.~Tam, and N.~Kottege, ``Autonomous obstacle legipulation with a
  hexapod robot,'' \emph{arXiv preprint arXiv:2011.06227}, 2020.

\bibitem{xu2020aanet}
H.~Xu and J.~Zhang, ``Aanet: Adaptive aggregation network for efficient stereo
  matching,'' in \emph{Proc. CVPR}, 2020, pp. 1959--1968.

\bibitem{zhang2019ga}
F.~Zhang, V.~Prisacariu, R.~Yang, and P.~H. Torr, ``Ga-net: Guided aggregation
  net for end-to-end stereo matching,'' in \emph{Proc. CVPR}, 2019, pp.
  185--194.

\bibitem{fujita2020fine}
H.~Fujita, M.~Itagaki, K.~Ichikawa, Y.~K. Hooi, K.~Kawano, and R.~Yamamoto,
  ``Fine-tuned pre-trained mask r-cnn models for surface object detection,''
  \emph{arXiv preprint arXiv:2010.11464}, 2020.

\bibitem{zhang2014loam}
J.~Zhang and S.~Singh, ``Loam: Lidar odometry and mapping in real-time,'' in
  \emph{Proc. RSS}, 2014, pp. 1--9.

\bibitem{schonberger2016structure}
J.~L. Schonberger and J.-M. Frahm, ``Structure-from-motion revisited,'' in
  \emph{Proc. CVPR}, 2016, pp. 4104--4113.

\bibitem{godard2019digging}
C.~Godard, O.~Mac~Aodha, M.~Firman, and G.~J. Brostow, ``Digging into
  self-supervised monocular depth estimation,'' in \emph{Proc. ICCV}, 2019, pp.
  3828--3838.

\bibitem{yin2018geonet}
Z.~Yin and J.~Shi, ``Geonet: Unsupervised learning of dense depth, optical flow
  and camera pose,'' in \emph{Proc. CVPR}, 2018, pp. 1983--1992.

\bibitem{tishchenko2020self}
I.~Tishchenko, S.~Lombardi, M.~R. Oswald, and M.~Pollefeys, ``Self-supervised
  learning of non-rigid residual flow and ego-motion,'' in \emph{Proc. 3DV},
  2020, pp. 150--159.

\bibitem{eigen2014depth}
D.~Eigen, C.~Puhrsch, and R.~Fergus, ``Depth map prediction from a single image
  using a multi-scale deep network,'' in \emph{Proc. NIPS}, 2014, pp.
  2366--2374.

\bibitem{zhou2017unsupervised}
T.~Zhou, M.~Brown, N.~Snavely, and D.~G. Lowe, ``Unsupervised learning of depth
  and ego-motion from video,'' in \emph{Proc. CVPR}, 2017, pp. 1851--1858.

\bibitem{irani1996parallax}
M.~Irani and P.~Anandan, ``Parallax geometry of pairs of points for 3d scene
  analysis,'' in \emph{Proc. ECCV}, 1996, pp. 17--30.

\bibitem{irani1997recovery}
M.~Irani, B.~Rousso, and S.~Peleg, ``Recovery of ego-motion using region
  alignment,'' \emph{IEEE TPAMI}, vol.~19, no.~3, pp. 268--272, 1997.

\bibitem{kumar1994direct}
R.~Kumar, P.~Anandan, and K.~Hanna, ``Direct recovery of shape from multiple
  views: A parallax based approach,'' in \emph{Proc. ICPR}, 1994, pp. 685--688.

\bibitem{shashua1994relative}
A.~Shashua and N.~Navab, ``Relative affine structure: Theory and application to
  3d reconstruction from perspective views,'' in \emph{Proc. CVPR}, 1994, pp.
  483--489.

\bibitem{irani2002direct}
M.~Irani, P.~Anandan, and M.~Cohen, ``Direct recovery of planar-parallax from
  multiple frames,'' \emph{IEEE TPAMI}, vol.~24, no.~11, pp. 1528--1534, 2002.

\bibitem{sun2020scalability}
P.~Sun, H.~Kretzschmar, X.~Dotiwalla, A.~Chouard, V.~Patnaik, P.~Tsui, J.~Guo,
  Y.~Zhou, Y.~Chai, B.~Caine, V.~Vasudevan, W.~Han, J.~Ngiam, H.~Zhao,
  A.~Timofeev, S.~Ettinger, M.~Krivokon, A.~Gao, A.~Joshi, Y.~Zhang, J.~Shlens,
  Z.~Chen, and D.~Anguelov, ``Scalability in perception for autonomous driving:
  Waymo open dataset,'' in \emph{Proc. CVPR}, 2020, pp. 2446--2454.

\bibitem{sawhney19943d}
H.~S. Sawhney, ``3d geometry from planar parallax,'' in \emph{Proc. CVPR},
  1994, pp. 929--934.

\bibitem{chaney2019learning}
K.~Chaney, A.~Z. Zhu, and K.~Daniilidis, ``Learning event-based height from
  plane and parallax,'' in \emph{Proc. IROS}, 2019, p. 3690–3696.

\bibitem{xing2022joint}
H.~Xing, Y.~Cao, M.~Biber, M.~Zhou, and D.~Burschka, ``Joint prediction of
  monocular depth and structure using planar and parallax geometry,''
  \emph{PR}, vol. 130, p. 108806, 2022.

\bibitem{fan2017point}
H.~Fan, H.~Su, and L.~J. Guibas, ``A point set generation network for 3d object
  reconstruction from a single image,'' in \emph{Proc. CVPR}, 2017, pp.
  605--613.

\bibitem{choy20163d}
C.~B. Choy, D.~Xu, J.~Gwak, K.~Chen, and S.~Savarese, ``3d-r2n2: A unified
  approach for single and multi-view 3d object reconstruction,'' in \emph{Proc.
  ECCV}, 2016, pp. 628--644.

\bibitem{wang2018pixel2mesh}
N.~Wang, Y.~Zhang, Z.~Li, Y.~Fu, W.~Liu, and Y.-G. Jiang, ``Pixel2mesh:
  Generating 3d mesh models from single rgb images,'' in \emph{Proc. ECCV},
  2018, pp. 52--67.

\bibitem{saito2019pifu}
S.~Saito, Z.~Huang, R.~Natsume, S.~Morishima, A.~Kanazawa, and H.~Li, ``Pifu:
  Pixel-aligned implicit function for high-resolution clothed human
  digitization,'' in \emph{Proc. ICCV}, 2019, pp. 2304--2314.

\bibitem{garg2016unsupervised}
R.~Garg, V.~K. Bg, G.~Carneiro, and I.~Reid, ``Unsupervised cnn for single view
  depth estimation: Geometry to the rescue,'' in \emph{Proc. ECCV}, 2016, pp.
  740--756.

\bibitem{godard2017unsupervised}
C.~Godard, O.~Mac~Aodha, and G.~J. Brostow, ``Unsupervised monocular depth
  estimation with left-right consistency,'' in \emph{Proc. CVPR}, 2017, pp.
  270--279.

\bibitem{zbontar2015computing}
J.~Zbontar and Y.~LeCun, ``Computing the stereo matching cost with a
  convolutional neural network,'' in \emph{Proc. CVPR}, 2015, pp. 1592--1599.

\bibitem{kendall2017end}
A.~Kendall, H.~Martirosyan, S.~Dasgupta, P.~Henry, R.~Kennedy, A.~Bachrach, and
  A.~Bry, ``End-to-end learning of geometry and context for deep stereo
  regression,'' in \emph{Proc. ICCV}, 2017, pp. 66--75.

\bibitem{chang2018pyramid}
J.-R. Chang and Y.-S. Chen, ``Pyramid stereo matching network,'' in \emph{Proc.
  CVPR}, 2018, pp. 5410--5418.

\bibitem{guo2019group}
X.~Guo, K.~Yang, W.~Yang, X.~Wang, and H.~Li, ``Group-wise correlation stereo
  network,'' in \emph{Proc. CVPR}, 2019, pp. 3273--3282.

\bibitem{vaswani2017attention}
A.~Vaswani, N.~Shazeer, N.~Parmar, J.~Uszkoreit, L.~Jones, A.~N. Gomez,
  L.~Kaiser, and I.~Polosukhin, ``Attention is all you need,'' in \emph{Proc.
  NIPS}, 2017, pp. 5998--6008.

\bibitem{han2020survey}
K.~Han, Y.~Wang, H.~Chen, X.~Chen, J.~Guo, Z.~Liu, Y.~Tang, A.~Xiao, C.~Xu,
  Y.~Xu \emph{et~al.}, ``A survey on visual transformer,'' \emph{IEEE TPMAI},
  vol.~45, no.~1, pp. 87--110, 2023.

\bibitem{dosovitskiy2021image}
A.~Dosovitskiy, L.~Beyer, A.~Kolesnikov, D.~Weissenborn, X.~Zhai,
  T.~Unterthiner, M.~Dehghani, M.~Minderer, G.~Heigold, S.~Gelly \emph{et~al.},
  ``An image is worth 16x16 words: Transformers for image recognition at
  scale,'' in \emph{Proc. ICLR}, 2021, pp. 1--9.

\bibitem{carion2020end}
N.~Carion, F.~Massa, G.~Synnaeve, N.~Usunier, A.~Kirillov, and S.~Zagoruyko,
  ``End-to-end object detection with transformers,'' in \emph{Proc. ECCV},
  2020, pp. 213--229.

\bibitem{zhu2021deformable}
X.~Zhu, W.~Su, L.~Lu, B.~Li, X.~Wang, and J.~Dai, ``Deformable detr: Deformable
  transformers for end-to-end object detection,'' in \emph{Proc. ICLR}, 2021,
  pp. 1--9.

\bibitem{parmar2018image}
N.~Parmar, A.~Vaswani, J.~Uszkoreit, L.~Kaiser, N.~Shazeer, A.~Ku, and D.~Tran,
  ``Image transformer,'' in \emph{Proc. ICML}, 2018, pp. 4055--4064.

\bibitem{liu2021Swin}
Z.~Liu, Y.~Lin, Y.~Cao, H.~Hu, Y.~Wei, Z.~Zhang, S.~Lin, and B.~Guo, ``Swin
  transformer: Hierarchical vision transformer using shifted windows,'' in
  \emph{Proc. ICCV}, 2021, pp. 10\,012--10\,022.

\bibitem{ho2019axial}
J.~Ho, N.~Kalchbrenner, D.~Weissenborn, and T.~Salimans, ``Axial attention in
  multidimensional transformers,'' \emph{arXiv preprint arXiv:1912.12180},
  2019.

\bibitem{wang2020parallax}
L.~Wang, Y.~Guo, Y.~Wang, Z.~Liang, Z.~Lin, J.~Yang, and W.~An, ``Parallax
  attention for unsupervised stereo correspondence learning,'' \emph{IEEE
  TPAMI}, vol.~44, no.~4, pp. 2108--2125, 2020.

\bibitem{li2020revisiting}
Z.~Li, X.~Liu, N.~Drenkow, A.~Ding, F.~X. Creighton, R.~H. Taylor, and
  M.~Unberath, ``Revisiting stereo depth estimation from a sequence-to-sequence
  perspective with transformers,'' in \emph{Proc. ICCV}, 2021, pp. 6197--6206.

\bibitem{guizilini20203d}
V.~Guizilini, R.~Ambrus, S.~Pillai, A.~Raventos, and A.~Gaidon, ``{3D} packing
  for self-supervised monocular depth estimation,'' in \emph{Proc. CVPR}, 2020,
  pp. 2485--2494.

\bibitem{chen2020exploring}
X.~Chen and K.~He, ``Exploring simple siamese representation learning,'' in
  \emph{Proc. CVPR}, 2021, pp. 15\,750--15\,758.

\bibitem{wang2019learning}
L.~Wang, Y.~Wang, Z.~Liang, Z.~Lin, J.~Yang, W.~An, and Y.~Guo, ``Learning
  parallax attention for stereo image super-resolution,'' in \emph{Proc. CVPR},
  2019, pp. 12\,250--12\,259.

\bibitem{yu2016multi}
F.~Yu and V.~Koltun, ``Multi-scale context aggregation by dilated
  convolutions,'' \emph{Proc. ICLR}, pp. 1--9, 2016.

\bibitem{wang2004image}
Z.~Wang, A.~C. Bovik, H.~R. Sheikh, and E.~P. Simoncelli, ``Image quality
  assessment: from error visibility to structural similarity,'' \emph{IEEE
  TIP}, vol.~13, no.~4, pp. 600--612, 2004.

\bibitem{bian2019unsupervised}
J.-W. Bian, Z.~Li, N.~Wang, H.~Zhan, C.~Shen, M.-M. Cheng, and I.~Reid,
  ``Unsupervised scale-consistent depth and ego-motion learning from monocular
  video,'' in \emph{Proc. NeurIPS}, 2019, pp. 35--45.

\bibitem{meister2018unflow}
S.~Meister, J.~Hur, and S.~Roth, ``Unflow: Unsupervised learning of optical
  flow with a bidirectional census loss,'' in \emph{Proc. AAAI}, 2018, pp.
  7251--7259.

\bibitem{jonschkowski2020matters}
R.~Jonschkowski, A.~Stone, J.~T. Barron, A.~Gordon, K.~Konolige, and
  A.~Angelova, ``What matters in unsupervised optical flow,'' in \emph{Proc.
  ECCV}, 2020, pp. 557--572.

\bibitem{fischler1981random}
M.~A. Fischler and R.~C. Bolles, ``Random sample consensus: a paradigm for
  model fitting with applications to image analysis and automated
  cartography,'' \emph{Communications of the ACM}, vol.~24, no.~6, pp.
  381--395, 1981.

\bibitem{paszke2019pytorch}
A.~Paszke, S.~Gross, F.~Massa, A.~Lerer, J.~Bradbury, G.~Chanan, T.~Killeen,
  Z.~Lin, N.~Gimelshein, L.~Antiga \emph{et~al.}, ``Pytorch: An imperative
  style, high-performance deep learning library,'' in \emph{Proc. NeurIPS},
  2019, pp. 8026--8037.

\bibitem{kingma2015adam}
D.~P. Kingma and J.~Ba, ``Adam: A method for stochastic optimization,'' in
  \emph{Proc. ICLR}, 2015, pp. 1--9.

\bibitem{ranftl2021vision}
R.~Ranftl, A.~Bochkovskiy, and V.~Koltun, ``Vision transformers for dense
  prediction,'' in \emph{Proc. CVPR}, 2021, pp. 12\,179--12\,188.

\bibitem{he2016deep}
K.~He, X.~Zhang, S.~Ren, and J.~Sun, ``Deep residual learning for image
  recognition,'' in \emph{Proc. CVPR}, 2016, pp. 770--778.

\end{thebibliography}

\end{document}